\documentclass{article} % For LaTeX2e
\usepackage{colm2024}

\usepackage{microtype}
\usepackage{hyperref}
\usepackage{url}
\usepackage{booktabs}

\usepackage{xspace}
\usepackage{xcolor}
\usepackage{wrapfig}
\usepackage{caption}
\usepackage[T1]{fontenc}
\usepackage{newfloat}
\usepackage{graphicx}
\usepackage{booktabs}
\usepackage{multirow}
\usepackage{enumitem}
\usepackage{adjustbox}
\usepackage[most]{tcolorbox}

\tcbset{colframe=black, colback=white, breakable, enhanced}
\newtcolorbox[auto counter]{prompt}[2][]{
  enhanced, breakable,
  colframe=darkgray, colback=white,
  left=0.5em, right=0.5em, toptitle=0.15em,
  label=#1,
  title={Prompt \thetcbcounter: #2}
}
\makeatletter
\newcommand{\tcbseparator}{
  \\
  \begin{tikzpicture}
  \path[use as bounding box] (0,0) -- (\linewidth,0);
  \draw[color=darkgray,dashed,dash phase=2pt]
        (0-\kvtcb@leftlower-\kvtcb@boxsep,0)--
        (\linewidth+\kvtcb@rightlower+\kvtcb@boxsep,0);
  \end{tikzpicture}
  \\[0.5em]
  }
\makeatother

\newcommand{\method}{HtT\xspace}

\title{Large Language Models can Learn Rules}

\author{Zhaocheng Zhu\textsuperscript{1, 2, 3,}\thanks{Work done during an internship at Google.} \quad Yuan Xue\textsuperscript{1} \quad Xinyun Chen\textsuperscript{1} \quad Denny Zhou\textsuperscript{1} \quad Jian Tang\textsuperscript{2, 4, 6} \\
\textbf{Dale Schuurmans\textsuperscript{1, 5, 6} \quad Hanjun Dai\textsuperscript{1}} \\
\textsuperscript{1} Google \quad \textsuperscript{2} Mila - Qu\'ebec AI Insitute \quad \textsuperscript{3} Universit\'e de Montr\'eal \\
\textsuperscript{4} HEC Montr\'eal \quad \textsuperscript{5} University of Alberta \quad \textsuperscript{6} CIFAR AI Chair \\
\texttt{zhuzhaoc@mila.quebec, hadai@google.com}
}

\renewcommand{\headrulewidth}{1pt}
\fancypagestyle{firstpage}{
  \lhead{\begin{picture}(0,0)\put(0,-3){\includegraphics[width=0.25\linewidth]{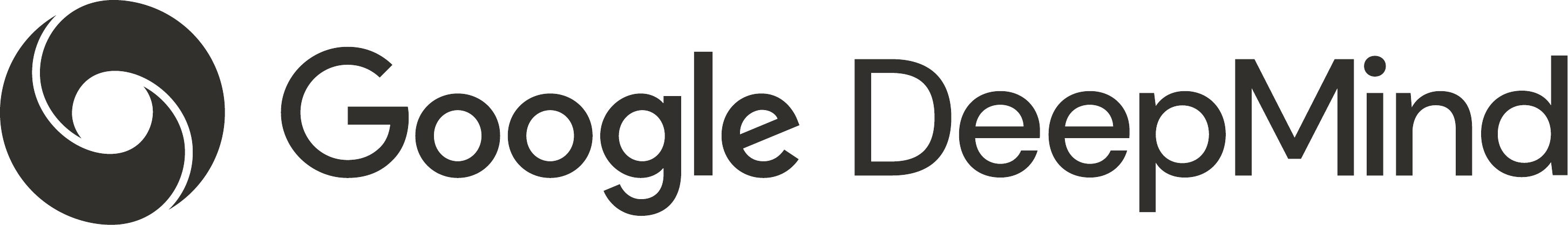}}\end{picture}}
}

\colmfinalcopy % Uncomment for camera-ready version, but NOT for submission.

\begin{document}

\maketitle

\thispagestyle{firstpage}
\begin{abstract}

When prompted with a few examples and intermediate steps, large language models (LLMs) have demonstrated impressive performance in various reasoning tasks. However, prompting methods that rely on implicit knowledge in an LLM often generate incorrect answers when the implicit knowledge is wrong or inconsistent with the task. To tackle this problem, we present Hypotheses-to-Theories (\method)\footnote{Code is available at \url{https://github.com/google-deepmind/llms_can_learn_rules}}, a framework that learns a rule library for reasoning with LLMs. \method contains two stages, an induction stage and a deduction stage. In the induction stage, an LLM is first asked to generate and verify rules over a set of training examples. Rules that appear and lead to correct answers sufficiently often are collected to form a rule library. In the deduction stage, the LLM is then prompted to employ the learned rule library to perform reasoning to answer test questions. Experiments on relational reasoning, numerical reasoning and concept learning problems show that \method improves existing prompting methods, with an absolute gain of 10-30\% in accuracy. The learned rules are also transferable to different models and to different forms of the same problem.

\end{abstract}

\begin{figure*}[!h]
    \vspace{-1em}
    \centering
    \includegraphics[width=\textwidth]{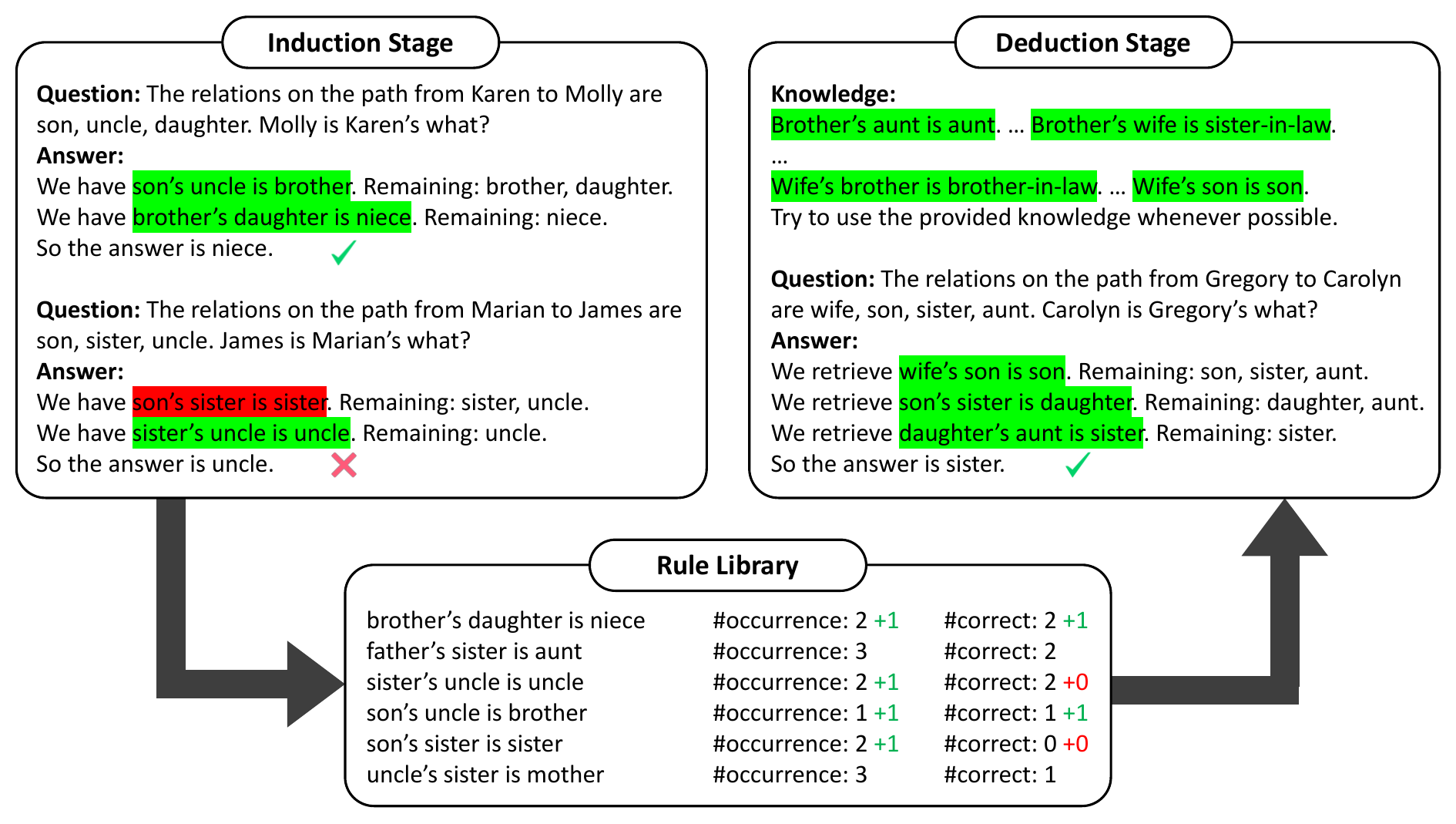}
    \vspace{-1.5em}
    \caption{An example of Hypotheses-to-Theories applied to chain-of-thought for the relational reasoning problem. Few-shot examples are omitted for brevity. The induction stage uses CoT to generate rules and verify them on the training samples. Rules are then collected and filtered to form the rule library. The deduction stage augments CoT with knowledge from the rule library. Correct and incorrect rules are marked with green and red respectively.}
    \label{fig:prompt}
    \vspace{-1em}
\end{figure*}

\section{Introduction}

Coinciding with their tremendous growth in scale, large language models (LLMs)~\citep[][\textit{inter alia}]{brown2020language, chowdhery2022palm, openai2023gpt4, anil2023palm, team2023gemini} have demonstrated emergent capabilities across a wide range of reasoning tasks~\citep{wei2022emergent, bubeck2023sparks}, including program synthesis, arithmetic reasoning, symbolic reasoning and commonsense reasoning. Importantly, these abilities are commonly elicited by advanced prompting techniques~\citep{wei2022chain, zhou2023least, khot2023decomposed} that teach an LLM to decompose a complex problem into simple steps and perform reasoning step by step based on a small set of in-context examples.

For many reasoning problems, decomposing and conducting reasoning steps are not sufficient to solve the problem, since one needs domain-specific knowledge to generate correct steps. For instance, when inferring the relationship between two people in a family tree (Figure~\ref{fig:prompt}), an LLM should know the basic rules\footnote{In this paper, rules to refer to intermediate steps that are reusable across different samples. This is slightly different from the definition of rules in formal logic.} to merge family relations. However, LLMs are often prone to errors in generating such rules~\citep{zheng2023does, zhang2023language}, especially when the task deviates from requiring conventional knowledge (e.g.\ arithmetic in a non-decimal system)~\citep{tang2023large, wu2023reasoning}. To solve this challenge, one should find a way to equip LLMs with those domain-specific rules. While it is always possible to curate a dataset and inject required rules into an LLM via supervised finetuning~\citep{wang2021kepler, talmor2020leap, nye2021show}, we are interested in a generic solution that enables LLMs to automatically discover rules from standard datasets without rule annotation.

This paper proposes such a solution for LLMs to learn a library of textual rules and apply the rule library to solve new samples. Our framework, dubbed Hypotheses-to-Theories (\method), consists of an induction stage and a deduction stage, akin to training and test stages of neural networks respectively. In the induction stage, an LLM is asked to generate rules for each example in the training set. The rules are verified by comparing the prediction of the LLM with the ground truth answer. We then filter the rules based on their number of occurrence and frequency of association with correct answers to construct the rule library for the deduction stage. In the deduction stage, the LLM is then asked to apply the learned rules to solve the reasoning problem, thereby reducing the chance of generating incorrect rules. To reduce the effort required for prompt engineering, we propose a technique called \emph{induction from deduction}, which fuses the rule generation and verification steps into a single deduction-like step. In this way, the prompts for both stages can be easily derived from existing few-shot prompting methods, such as chain-of-thought or least-to-most prompting.

Empirically, we verify the effectiveness of \method with GPT-3.5 and GPT-4~\citep{openai2023gpt4} on the CLUTRR~\citep{sinha2019clutrr}, Arithmetic~\cite{wu2023reasoning} and List Functions~\citep{rule2020child} datasets, which correspond to relational reasoning, numerical reasoning and concept learning respectively. Experiments show that \method consistently improves over baseline prompting methods across the models and datasets considered, with an absolute gain of 10-30\% in most cases. Moreover, the learned rules can be directly transferred to the textual version of CLUTRR, providing a practical advantage over previous reasoning approaches. Besides, We conduct extensive ablation studies to understand the properties of \method, finding that the performance gain arises primarily from a reduction in the number of incorrect rules due to the use of the learned rules. We also observe a log-linear scaling law between accuracy and the number of training examples on all three datasets.
\vspace{-0.2em}
\section{Preliminaries}

Intuitively, reasoning is the process of applying sound logical steps to draw conclusions from existing information. At the core of reasoning are facts and rules. Facts are pieces of information that describe the current state, while rules are functions that transform facts into new facts under certain conditions. Take the famous syllogism as an example.
\begin{quote}
    \centering
    \vspace{-0.05em}
    \emph{Socrates is a man. \quad All men are mortal. \quad Therefore, Socrates is mortal.}
    \vspace{-0.05em}
\end{quote}
Here both \emph{``Socrates is mortal''} and \emph{``Socrates is a man''} are facts. \emph{``All men are mortal''} could be thought of as a rule, with the understanding that facts and rules are relative concepts intuitively.  That is, a rule can also be viewed as a fact when coupled with a high-order rule, e.g.\ \emph{``All men are mortal''} is a fact w.r.t. the rule \emph{``If all men are animals, then all men are mortal''}.

\textbf{Deductive Reasoning.}
Deductive reasoning aims to derive new facts based on known facts and rules. The above syllogism is a basic deductive step. When there are multiple facts and rules, one needs to match rules against facts, which is known as \emph{unification}. Generally, there are many ways to apply unification at each step, and a planning algorithm is required to find the optimal ordering of steps, e.g.\ forward or backward chaining~\citep{stuart2016artificial}.

For LLMs, most prompting techniques are designed to elicit deductive reasoning under the assumption of a greedy plan, such as chain-of-thought~\citep{wei2022chain} and least-to-most~\citep{zhou2023least} prompting. We categorize these as implicit reasoning methods since they rely on implicit rules stored in the LLM. By contrast, explicit reasoning methods such as Selection-Inference~\citep{creswell2023selection} and LAMBADA~\citep{kazemi2023lambada} operate on given facts and rules, focusing on searching an optimal deduction order.

\textbf{Inductive Reasoning.}
Inductive reasoning focuses on deriving general rules from observed facts. For example, if we know \emph{``Scorates is a man''} and \emph{``Socrates is mortal''} and the same facts hold for \emph{Aristotle}, we might hypothesize a rule \emph{``All men are mortal''}. While many rules can be induced from given facts, some rules are overly specific to the given facts, e.g.\ \emph{``If Scorates is a man, Socrates is mortal''} can only be applied to \emph{Scorates}. To make rules useful for future prediction, they should have a large set of supporting examples (i.e.\ coverage) while also predicting new facts correctly (i.e.\ confidence)~\citep{agrawal1994fast, galarraga2013amie}.

In machine learning, inductive reasoning is usually not a standalone task, since it is challenging to annotate and evaluate rules. Instead, inductiveness is studied as a desired property for deductive reasoning models to generalize to new facts. Examples include inductive logic programming (ILP)~\citep{muggleton1994inductive} that requires models to generalize to new combinations of elements, and knowledge graph reasoning~\citep{teru2020inductive, galkin2024towards} that requires models to generalize to graphs of new entities and relations.
\section{Hypotheses-to-Theories: Learning a Rule Library for Reasoning}

For many reasoning problems, the problem statements only contain the necessary facts within the context, but the rules are not explicitly stated. For instance, when an LLM is asked infer the relationship between two people, it is not given any kinship rules. An LLM pretrained on a massive corpus is able to recall certain commonsense rules from its parameters~\citep{petroni2019language, roberts2020much}, but due to the implicit nature of this process, it may often generate incorrect rules when solving reasoning problems. Our manual analysis indicates that such incorrect rules constitute 65\% and 81\% of the errors made by CoT on CLUTRR and base-16 Arithmetic respectively (Figure~\ref{fig:error_case}).

To solve the above challenge, we propose Hypotheses-to-Theories (\method), a framework that learns a textual rule library from training examples and explicitly uses the rule library to solve test samples. \method consists of an induction stage and a deduction stage, both implemented by few-shot prompting. In the induction stage, rules are generated and verified on a set of question-answer examples. The rules are then collected and filtered to form a library. In the deduction stage, we prompt the model to explicitly retrieve rules from the rule library to answer test questions. The two stages are similar to training and test stages of neural networks, except that we learn textual rules instead of model parameters.

\vspace{-0.2em}
\subsection{Induction Stage: Learning A Rule Library By Generation and Verification}

The induction stage aims to learn rules from training examples without rule annotation. For each training example (a question-answer pair), we ask an LLM to generate rules for answering the question. We extract rules from the output of the LLM with regular expressions, assuming the LLM follows the template of few-shot exemplars. Note that these rules can be either correct or incorrect. While we cannot judge these rules without golden rules, we can verify them by comparing the prediction of these rules against the ground truth answer. The insight is that if the model predicts the correct answer, it is very likely that all these rules are correct. Conversely, if the model predicts a wrong answer, it is very likely that at least one of the rules is incorrect. Due to the noisy nature of LLM reasoning, we collect rules and accuracy metrics from a reasonable number of training examples.

To filter the rules for the rule library, we follow the principles of rule mining~\citep{galarraga2013amie} and consider both coverage and confidence as criteria. The coverage of a rule tells how likely it will be reused, and the confidence of a rule indicates how likely it is correct. Specifically, we measure coverage based on the number of occurrence of each rule in the training set. Confidence is measured by the average accuracy of examples where a rule occurs. In practice, for each generated rule, we maintain two counters, number of occurrence and number of correct answers (Figure~\ref{fig:prompt} bottom). The confidence of each rule can be computed by dividing its number of correct answers by its number of occurrence. We fill the library with good rules exceeding a minimal coverage $k$ and a minimal confidence $p$.

\textbf{Induction from Deduction.}
The induction stage introduces two sub-problems, rule generation and verification. Recent works on induction~\citep{yang2024language, qiu2024phenomenal} use two separate prompts for generation and verification, i.e.\ a prompt for generating rules based on the question and a prompt for applying the rules to deduce the answer. While it is possible to use two prompts here as well, this doubles the prompt engineering effort and complicates comparisons with other methods. Moreover, the multi-step nature of reasoning problems makes it challenging to generate rules for later steps at the beginning. Hence we propose induction from deduction, which adapts a deductive reasoning prompt (e.g.\ CoT, LtM) for both rule generation and verification (Figure~\ref{fig:prompt} left). The key idea is to explicitly declare a rule whenever a deduction is performed. In this way, both induction and deduction stages use the same base prompt, which is directly comparable to the base prompting method.

\vspace{-0.2em}
\subsection{Deduction Stage: Explicit Reasoning with the Rule Library}
\label{sec:deduction}

In the deduction stage, we apply the rule library from the induction stage to solve test questions. Specifically, we prepend the rule library to a deductive reasoning prompt (e.g.\ CoT, LtM), and modify the exemplars to teach the LLM to retrieve rules from the library whenever it needs to generate a rule (Figure~\ref{fig:prompt} right). If all the rules required by a question are present in the library, the LLM should generate correct rules for each step without errors.

In practice, we find that even a strong LLM (e.g. GPT-4) somehow struggles to perform retrieval, especially when the library contains a large number of rules in an unstructured way. One workaround is to employ a pretrained passage retriever~\citep{karpukhin2020dense} and interleave retriever calls with LLM decoding~\citep{trivedi2023interleaving}. However, this introduces additional modules and makes the comparison against the base prompting method unfair. Here we develop a pure prompting solution: we organize the rule library into a hierarchy and use XML tags to explicitly refer to the clusters we want to retrieve from the hierarchy. More details of this XML tagging trick can be found in Appendix~\ref{app:implementation}.

\vspace{-0.2em}
\subsection{Discussion: Why Does \method Work? What Kind of Tasks Can It Solve?}

It might look surprising that simply prompting the LLM and verifying the predictions on sufficient training examples can give us a library of good rules. Here we discuss the key insights behind \method. While LLMs may occasionally generate incorrect rules, we conjecture they are able to produce correct rules on some examples with a non-trivial probability, similar to the assumption in \cite{wang2024chain}. With a sufficient set of training examples, we can extract most of the necessary rules for a problem class based on the coverage and confidence criteria. Since retrieving a rule is usually easier than generating the correct rule for an LLM, it will perform better on deductive reasoning when primed with a rule library.

Generally, \method has a similar scope of tasks to its base few-shot prompting method. To achieve substantial performance gain with \method, two constraints apply: (1) To fit the library into the prompt, the number of rules required to solve most problems of the task should be moderately small ($\leq$500 in our experiments), excluding tasks that cannot be efficiently described by rules (e.g.\ natural language entailment). (2) To successfully induce most rules, the base prompting method should have a reasonable performance on the training examples ($\geq$20\% accuracy in our experiments), excluding tasks that are difficult for existing few-shot prompting methods, such as tasks requiring planning abilities~\citep{saparov2023language}. \method does not impose constraints on the type of rules it learns. Our experiments show that \method can learn kinship rules, numerical rules or even free-form rules that transform a list.
\section{Experimental Evaluation}
\label{sec:experiment}

To evaluate \method, we apply it as an augmentation to existing few-shot prompting methods. We benchmark the performance of \method on relational reasoning and numerical reasoning that require multi-step reasoning and multiple rules, as well as concept learning that require a single complex rule. We also conduct ablation studies to thoroughly understand \method.

\vspace{-0.2em}
\subsection{Implementation Details}
\label{sec:implementation}

We evaluate \method and the baselines using two different LLMs, \texttt{gpt-3.5-turbo-0613} and \texttt{gpt-4-0613}. When the prompts exceed the 4k context length of \texttt{gpt-3.5-turbo-0613}, we use \texttt{gpt-3.5-turbo-16k-0613} instead. We use the default temperature of 1 for these models. Throughout the following, we will denote the two LLMs as GPT-3.5 and GPT-4 respectively. We further evaluate \method with \texttt{gemini-1.0-pro} in Appendix~\ref{app:gemini}.

On CLUTRR and Arithmetic, we perform the induction stage on 2,000 training examples for the proposed \method. When the training set contains fewer than 2,000 examples, we resample the training examples. For each task in List Functions, we perform the induction stage on 20 training-validation splits sampled from the original data. We search the hyperparameters of \method within the following grid: minimal coverage $k \in \{1, 2, 3\}$, minimal confidence $p \in \{0.1, 0.3, 0.5, 0.7, 0.9\}$. Due to the cost of LtM (3-5$\times$ compared to CoT), we induce rules and select the best hyperparameters based on CoT prompting, and only use LtM prompting for the deduction stage. This might slightly underestimate the performance of \method for LtM prompting. More details and hyperparameter configurations are in Appendix~\ref{app:implementation}.

\vspace{-0.2em}
\subsection{Relational Reasoning}

We evaluate \method on CLUTRR~\citep{sinha2019clutrr}, a relational reasoning dataset that queries the relationship between two family members in a family tree. CLUTRR comes in two forms: a symbolic version that only contains entities and their relationships, and a textual version that describes the relationships in a story. We evaluate \method on both versions. We generate dataset splits by uniformly sampling the standard splits from \cite{sinha2019clutrr}. We use 2,000 samples of 2 and 3 hop examples for training, and 200 samples of 2 to 10 hop examples for both validation and test. For reference, we reproduce EdgeTransformer~\citep{bergen2021systematic}, one of the best domain-specific models on CLUTRR, in the same setting.

Table~\ref{tab:clutrr} shows the results on CLUTRR. Here \method consistently improves both CoT and LtM prompting with both models by a margin of 11.1-16.4\% in average accuracy. Since induction is more challenging than deduction, we further evaluate GPT-3.5 with rules induced by GPT-4. Surprisingly, with rules from GPT-4, \method increases the performance of CoT on GPT-3.5 by 27.2\%, doubling the performance of CoT. Compared to the supervised baseline EdgeTransformer, the performance of 5-shot CoT + \method with GPT-4 is 7.5\% lower, which is reasonable since EdgeTransformer leverages forward chaining as a strong inductive bias and is specific to this problem. \method has two advantages over such domain-specific models: (1) \method does not require a predefined relation vocabulary. (2) The rules learned by \method can directly transfer to textual inputs. As shown in Table~\ref{tab:clutrr_text}, rules learned from the symbolic version can also improve the performance of GPT-4 on the textual version. The improvement is not significant for GPT-3.5, since it often produces errors other than incorrect rules.

\begin{table}[t]
    \vspace{-0.5em}
    \centering
    \footnotesize
    \caption{Results on the symbolic version of CLUTRR.}
    \vspace{-0.5em}
    \label{tab:clutrr}
    \begin{adjustbox}{max width=\textwidth}
        \begin{tabular}{llcccccccccc}
            \toprule
            \bf{Model} & \bf{Prompt} & \bf{2 hops} & \bf{3 hops} & \bf{4 hops} & \bf{5 hops} & \bf{6 hops} & \bf{7 hops} & \bf{8 hops} & \bf{9 hops} & \bf{10 hops} & \bf{Average} \\
            \midrule
            \multicolumn{2}{l}{EdgeTransformer} & 100.0 & 94.4 & 96.8 & 88.0 & 68.8 & 61.9 & 50.0 & 50.0 & 36.0 & 71.8 \\
            \midrule
            & 0-shot CoT & 50.0 & 22.2 & 12.9 & 8.0 & 12.5 & 9.5 & 10.0 & 3.8 & 4.0 & 14.8 \\
            \cmidrule{2-12}
            & 5-shot CoT & 0.0  & 27.8 & 45.2 & 36.0 & 18.8 & 19.0 & 16.7 & 11.5 & 16.0 & 21.2 \\
            & + HtT & 87.5 & 38.9 & 35.5 & 44.0 & 37.5 & 14.3 & 33.3 & 11.5 & 36.0 & \bf{37.6 (+16.4)} \\
            GPT-3.5 & + HtT (GPT-4) & 100.0 & 55.6 & 32.3 & 60.0 & 50.0 & 47.6 & 43.3 & 19.2 & 28.0 & \bf{48.4 (+27.2)} \\
            \cmidrule{2-12}
            & 5-shot LtM & 37.5 & 22.2 & 29.0 & 36.0 & 25.0 & 14.3 & 10.0 & 23.1 & 20.0 & 24.1 \\
            & + HtT & 100.0 & 33.3 & 32.3 & 48.0 & 31.3 & 33.3 & 23.3 & 34.6 & 28.0 & \bf{40.5 (+16.4)} \\
            & + HtT (GPT-4) & 75 & 44.4 & 41.9 & 52.0 & 37.5 & 33.3 & 23.3 & 19.2 & 16.0 & \bf{38.1 (+14.0)} \\
            \midrule
            & 0-shot CoT & 50.0 & 22.2 & 22.6 & 32.0 & 37.5 &38.1 & 33.3 & 46.2 & 16.0 & 33.1 \\
            \cmidrule{2-12}
            & 5-shot CoT & 50.0 & 55.6 & 71.0 & 80.0 & 50.0 & 52.4 & 30.0 & 46.2 & 20.0 & 50.6 \\
            GPT-4 & + HtT & 100.0 & 61.1 & 74.2 & 84.0 & 75.0 & 38.1 & 56.7 & 53.8 & 36.0 & \bf{64.3 (+13.7)} \\
            \cmidrule{2-12}
            & 5-shot LtM & 62.5 & 38.9 & 58.1 & 68.0 & 50.0 & 38.1 & 43.3 & 34.6 & 28.0 & 46.8 \\
            & + HtT & 100.0 & 55.6 & 77.4 & 80.0 & 75.0 & 38.1 & 36.7 & 38.5 & 20.0 & \bf{57.9 (+11.1)} \\
            \bottomrule
        \end{tabular}
    \end{adjustbox}
    \vspace{-0.5em}
\end{table}

\begin{table}[t]
    \centering
    \footnotesize
    \caption{Results on Arithmetic. Note that GPT-4 (5-shot CoT) has 99.1\% accuracy on base-10.}
    \vspace{-0.5em}
    \label{tab:arithmetic}
    \begin{adjustbox}{max width=\textwidth}
        \begin{tabular}{llcccccccccc}
            \toprule
            \multirow{2}{*}{\bf{Model}} & \multirow{2}{*}{\bf{Prompt}} & \multicolumn{3}{c}{\bf{Base-16}} & \multicolumn{3}{c}{\bf{Base-11}} & \multicolumn{3}{c}{\bf{Base-9}} & \multirow{2}{*}{\bf{Average}}\\
            & & \bf{2 digits} & \bf{3 digits} & \bf{4 digits} & \bf{2 digits} & \bf{3 digits} & \bf{4 digits} & \bf{2 digits} & \bf{3 digits} & \bf{4 digits} \\
            \midrule
            & 0-shot CoT & 30.6 & 10.5 & 0.0 & 5.6 & 5.3 & 0.0 & 11.1 & 10.5 & 0.0 & 8.2 \\
            \cmidrule{2-12}
            & 5-shot CoT & 83.3 & 34.2 & 11.5 & 5.6 & 2.6 & 0.0 & 25.0 & 13.2 & 11.5 & 20.8 \\
            & + HtT & 77.8 & 52.6 & 23.1 & 25.0 & 13.2 & 0.0 & 8.3 & 5.3 & 11.5 & \bf{24.1 (+3.3)} \\
            GPT-3.5 & + HtT (GPT-4) & 63.9 & 44.7 & 34.6 & 13.9 & 7.9 & 3.8 & 25.0 & 7.9 & 11.5 & \bf{23.7 (+2.9)} \\
            \cmidrule{2-12}
            & 5-shot LtM & 83.3 & 34.2 & 15.4 & 16.7 & 5.3 & 0.0 & 13.9 & 7.9 & 7.7 & 20.5 \\
            & + HtT & 80.6 & 39.5 & 26.9 & 16.7 & 2.6 & 3.8 & 19.4 & 5.3 & 3.8 & \bf{22.1 (+1.6)} \\
            & + HtT (GPT-4) & 72.2 & 31.6 & 30.8 & 47.2 & 15.8 & 11.5 & 44.4 & 21.1 & 15.4 & \bf{32.2 (+11.7)} \\
            \midrule
            & 0-shot CoT & 72.2 & 26.3 & 7.7 & 22.2 & 10.5 & 3.8 & 30.6 & 34.2 & 23.1 & 25.6 \\
            \cmidrule{2-12}
            & 5-shot CoT & 83.3 & 71.1 & 61.5 & 52.8 & 47.4 & 46.2 & 75.0 & 36.8 & 42.3 & 57.4 \\
            GPT-4 & + HtT & 100.0 & 94.7 & 84.6 & 88.9 & 71.1 & 46.2 & 86.1 & 68.4 & 65.4 & \bf{78.4 (+21.0)} \\
            \cmidrule{2-12}
            & 5-shot LtM & 88.9 & 81.6 & 61.5 & 52.8 & 47.4 & 30.8 & 52.8 & 31.6 & 11.5 & 51.0 \\
            & + HtT & 100.0 & 86.8 & 76.9 & 72.2 & 52.6 & 46.2 & 61.1 & 23.7 & 38.5 & \bf{62.0 (+11.0)} \\
            \bottomrule
        \end{tabular}
    \end{adjustbox}
    \vspace{-0.5em}
\end{table}

\vspace{-0.2em}
\subsection{Numerical Reasoning}

We use the Arithmetic dataset~\citep{wu2023reasoning} to evaluate the LLMs on numerical reasoning in non-decimal systems. This dataset contains summation problems over 2 to 4 digits in several base systems. Since the rules in a non-decimal system are mostly different from those in the decimal system, arithmetic is considered to be a counterfactual setting that requires an LLM to perform reasoning rather than reciting. To prepare the dataset for \method, we split it into training, validation and test. The training set contains 900 examples of 2 digit addition. Both the validation and test sets contain 100 examples of 2, 3 and 4 digit addition.

Table~\ref{tab:arithmetic} shows the results on Arithmetic. 0-shot CoT performs worst for both models in all base systems, because the LLMs with 0-shot CoT tend to convert non-decimal inputs to decimal, perform calculations, then revert to non-decimal, which is error prone due to the extra multiplications and divisions. For both CoT and LtM, \method consistently improves the accuracy of two models by a large margin. The performance gain is less significant for GPT-3.5, since it is worse at inducing correct rules and retrieving rules from the library. This can be fixed by using a stronger model to induce the rules and offloading the retrieval steps to a separate prompt like in LtM. We observe a large improvement of LtM + \method on GPT-3.5 with better rules from GPT-4, especially on base-11 and base-9 where GPT-3.5 struggles to induce correct rules. By contrast, there is no improvement for CoT + \method with the better rules, because GPT-3.5 has a strong tendency to rely on its own beliefs (i.e.\ mostly decimal rules) when prompted with CoT, similar to the observation in \cite{longpre2021entity}.

\vspace{-0.2em}
\subsection{Concept Learning}

To assess the potential of \method in learning complex rules, we further evaluate \method on the concept learning problem using List Functions~\citep{rule2020child}. This dataset aims to identify a function that maps each input list to its corresponding output list, with 250 tasks grouped into 3 subsets: simple operations over numbers between 0 and 9 (P1), simple operations over numbers between 0 and 99 (P2), difficult operations over numbers between 0 and 99 (P3). For each task, we split 32 input-output pairs into 16 training samples and 16 test samples. For \method, we further split the 16 training samples into 8 training samples and 8 validation samples to verify the rules based on the validation performance.

Table~\ref{tab:list_functions} shows the results on List Functions. Following \cite{qiu2024phenomenal}, we report both raw accuracy and task accuracy. Raw accuracy is the accuracy on test input-output pairs, while task accuracy is the ratio of tasks with all test input-output pairs correctly solved. \method consistently improves 4-shot CoT on both models, with a gain of 18.5-18.7\% in raw accuracy and 10.2-14.5\% in task accuracy. Surprisingly, GPT-4 can discover some very complex rules in List Functions, as shown in Table~\ref{tab:learned_rules}. With rules learned by GPT-4, the task accuracy of GPT-3.5 can be boosted to 34.4\%, doubling the performance of GPT-3.5 with 4-shot CoT. This suggests that GPT-3.5 can understand most rules learned by GPT-4, and the challenge of concept learning lies more in induction than deduction. We also observe that the performance of GPT-3.5 decreases drastically on tasks involving large numbers (P2) or difficult operations (P3). By contrast, the decrease in performance is less significant for GPT-4, indicating that GPT-4 is more robust across various levels of difficulty.

\begin{table}[t]
    \vspace{-0.5em}
    \centering
    \footnotesize
    \caption{Results on List Functions.}
    \vspace{-0.5em}
    \label{tab:list_functions}
    \begin{adjustbox}{max width=\textwidth}
        \begin{tabular}{llcccccccc}
            \toprule
            \multirow{2}{*}{\bf{Model}} & \multirow{2}{*}{\bf{Prompt}} & \multicolumn{4}{c}{\bf{Raw Accuracy}} & \multicolumn{4}{c}{\bf{Task Accuracy}} \\
            & & \bf{P1} & \bf{P2} & \bf{P3} & \bf{Average} & \bf{P1} & \bf{P2} & \bf{P3} & \bf{Average} \\
            \midrule
            & 0-shot CoT & 44.1 & 38.4 & 28.9 & 37.1 & 30.0 & 25.0 & 12.7 & 22.6 \\
            \cmidrule{2-10}
            \multirow{2}{*}{GPT-3.5} & 4-shot CoT & 32.8 & 32.2 & 19.3 & 28.1 & 22.5 & 15.0 & 8.0 & 15.2 \\
            & + HtT & 58.4 & 50.9 & 30.5 & \bf{46.6 (+18.5)} & 40.0 & 35.0 & 14.0 & \bf{29.7 (+14.5)} \\
            & + HtT (GPT-4) & 69.2 & 66.3 & 38.4 & \bf{58.0 (+29.9)} & 50.0 & 40.0 & 13.3 & \bf{34.4 (+19.2)} \\
            \midrule
            & 0-shot CoT & 69.3 & 56.6 & 48.3 & 58.1 & 51.3 & 45.0 & 26.0 & 40.8 \\
            \cmidrule{2-10}
            GPT-4 & 4-shot CoT & 67.3 & 60.9 & 43.9 & 57.4 & 53.8 & 55.0 & 29.3 & 46.0 \\
            & + HtT & 82.3 & 84.4 & 61.5 & \bf{76.1 (+18.7)} & 61.3 & 70.0 & 37.3 & \bf{56.2 (+10.2)} \\
            \bottomrule
        \end{tabular}
    \end{adjustbox}
    \vspace{-0.5em}
\end{table}

\begin{table}[t]
    \centering
    \footnotesize
    \caption{Examples of complex rules learned by GPT-4 on List Functions.}
    \vspace{-0.5em}
    \label{tab:learned_rules}
    \begin{adjustbox}{max width=\textwidth}
        \begin{tabular}{lp{17em}p{18em}}
            \toprule
            \bf{Task ID} & \bf{Ground Truth} & \bf{Top Learned Rule} \\
            \midrule
            c085 & remove all but element N + 1, N = element 1. & return a list with the element that corresponds to the first number in the list, where indexing starts at 1. \\
            \midrule
            c191 & repeat each element N times, where N is its tens digit, in order of appearance. & list each element as many times as its tens digit. \\
            \bottomrule
        \end{tabular}
    \end{adjustbox}
    \vspace{-0.5em}
\end{table}

\vspace{-0.2em}
\subsection{Ablation Studies}
\label{sec:ablation}

We conduct ablation studies with GPT-4, since GPT-3.5 sometimes struggles to induce and retrieve rules. Full results of these experiments can be found in Appendix~\ref{app:experiment}.

\begin{table}[t]
    \vspace{-0.5em}
    \centering
    \footnotesize
    \begin{minipage}[t]{0.48\textwidth}
        \centering
        \caption{Results on the textual CLUTRR with rules learned on the symbolic CLUTRR.}
        \vspace{-0.5em}
        \begin{tabular}{llc}
            \toprule
            \bf{Model} & \bf{Prompt} & \bf{Accuracy} \\
            \midrule 
            \multirow{2}{*}{GPT-3.5}
            & 5-shot CoT & 16.0 \\
            & + \method & 16.3 (+0.3) \\
            \midrule
            \multirow{2}{*}{GPT-4}
            & 5-shot CoT & 48.7 \\
            & + \method & \bf{59.1 (+10.4)} \\
            \bottomrule
        \end{tabular}
        \label{tab:clutrr_text}
    \end{minipage}
    \hfill
    \begin{minipage}[t]{0.5\textwidth}
        \centering
        \caption{Comparison of random rules and learned rules on CLUTRR and Arithmetic.}
        \vspace{-0.5em}
        \begin{adjustbox}{max width=\textwidth}
            \begin{tabular}{lcc}
                \toprule
                \bf{Prompt} & \bf{CLUTRR} & \bf{Arithmetic} \\
                \midrule
                5-shot CoT & 50.6 & 57.4 \\
                \midrule
                + random rules & 9.9 (-40.7) & 23.7 (-33.7)\\
                + \method & \bf{64.3} (+13.7) & \bf{78.4} (+21.0) \\
                \bottomrule
            \end{tabular}
        \end{adjustbox}
        \label{tab:ablation}
    \end{minipage}
\end{table}

\begin{figure}[t]
    \vspace{-0.5em}
    \centering
    \includegraphics[width=0.9\textwidth]{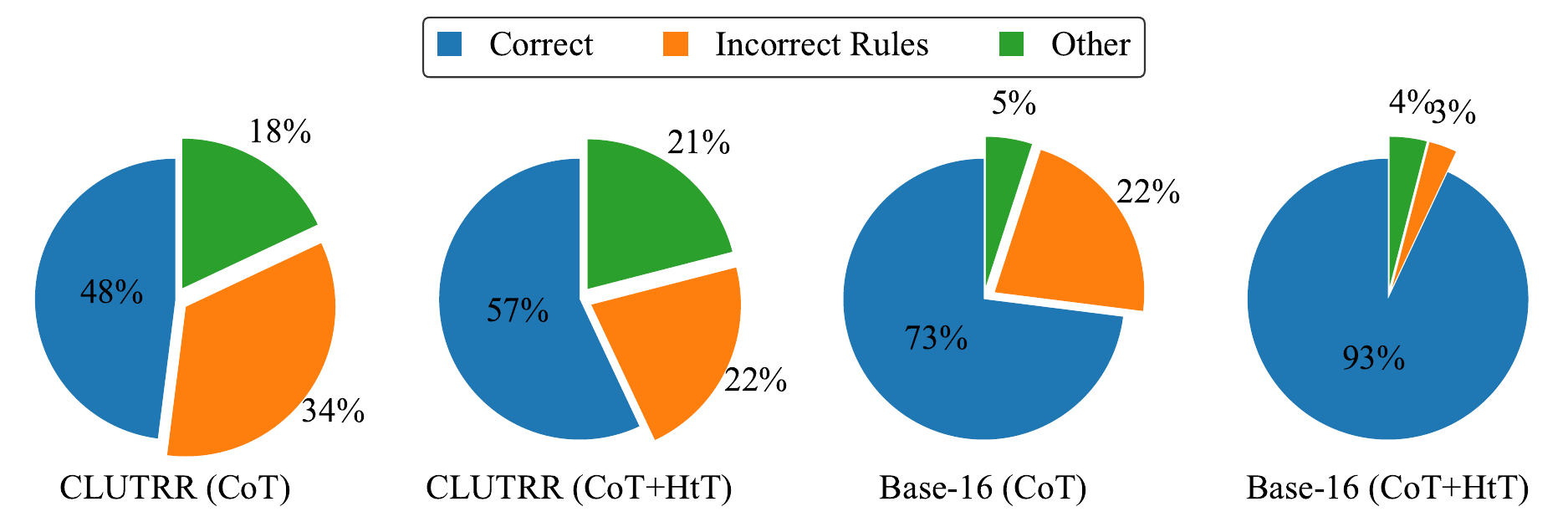}
    \vspace{-0.5em}
    \caption{Statistics of different error cases on CLUTRR and Arithmetic (base-16).}
    \label{fig:error_case}
    \vspace{-0.5em}
\end{figure}

\textbf{Does \method reduce the occurrence of incorrect rules?}
Since an LLM generates free-form text to solve a problem, there can be multiple reasons for failure~\citep{zheng2023does}. While \method boosts the overall performance on reasoning problems, it is not clear if the gain comes from less incorrect rules. Here we manually analyze the predictions of CoT and CoT + \method on 100 test examples from CLUTRR and Arithmetic (base-16), and classify the predictions into 3 categories: correct, incorrect rules and other. Figure~\ref{fig:error_case} plots the distribution of error cases. We can see that most performance gain of \method comes from reduction in incorrect rules.

\textbf{Do the learned rules just hint the model about the rule space?}
A previous study~\citep{min2022rethinking} found that random labels perform similarly to gold labels in in-context learning. If that was the case for our problems, we could just generate random rules and do not have to learn the rule library. To answer this question, we replace the conclusions in the learned rules with random answers, e.g.\ changing 5 + A = E to 5 + A = 7 in base-16. Table~\ref{tab:ablation} shows that random rules significantly hurt performance, indicating the necessity of learned rules in \method. We conjecture that the contrary observation is because \cite{min2022rethinking} studied simple classification problems, whereas we are dealing with multi-step reasoning problems.

\begin{figure}[t]
    \centering
    \vspace{-0.5em}
    \includegraphics[width=\textwidth]{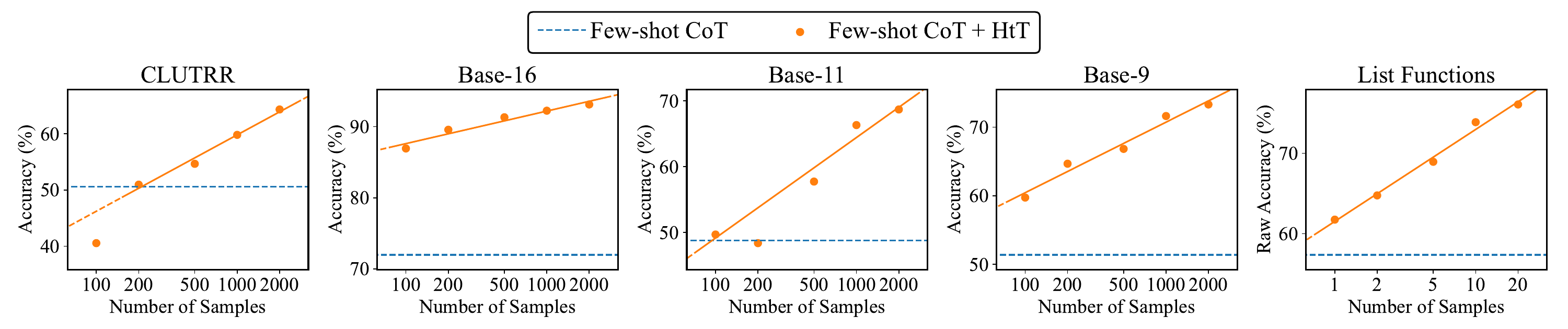}
    \vspace{-5mm}
    \caption{Performance of \method w.r.t. the number of samples in the induction stage.}
    \label{fig:training_samples}
    \vspace{-0.5em}
\end{figure}

\begin{figure}[t]
    \centering
    \vspace{-0.5em}
    \includegraphics[width=0.66\textwidth]{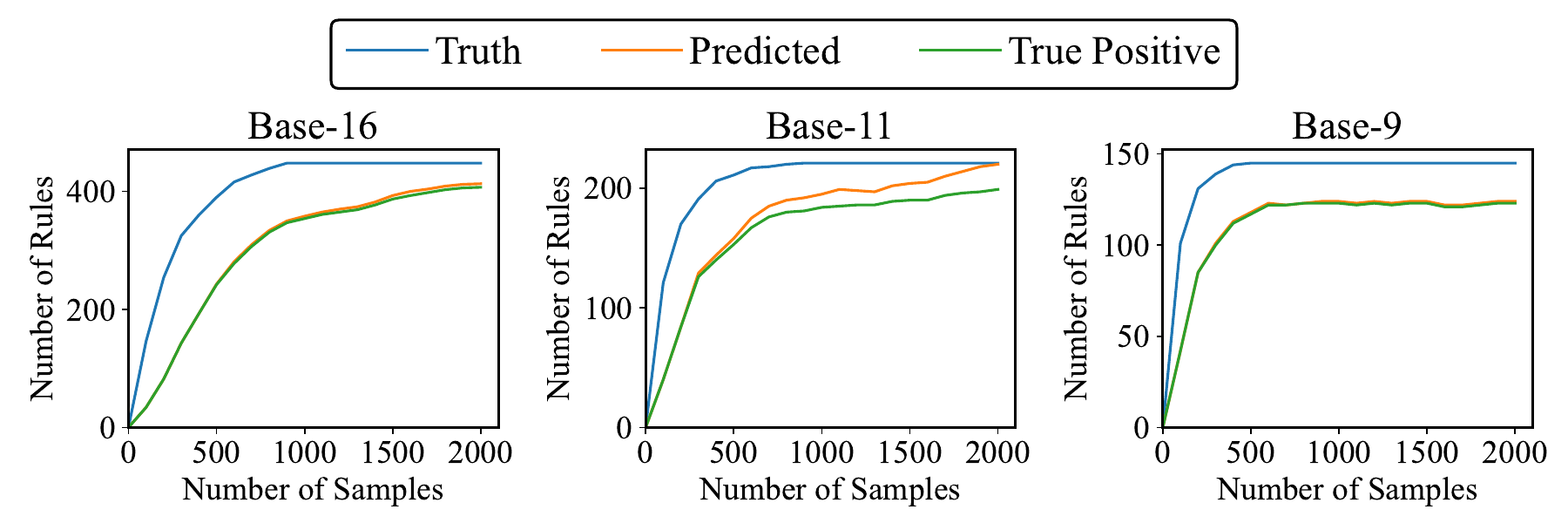}
    \vspace{-0.5em}
    \caption{Number of rules discovered by \method in the induction stage.}
    \label{fig:number_of_rules}
    \vspace{-0.5em}
\end{figure}

\textbf{How many samples does \method need for the induction stage?}
One may be curious about how \method scales with the number of samples and what is the minimal number of examples required. Here we conduct experiments with different numbers of examples for the induction stage. As shown in Figure~\ref{fig:training_samples}, there is a log-linear trend between performance and the number of examples, consistent with the scaling law for supervised learning~\citep{kaplan2020scaling}. The minimal number of examples varies across datasets. CLUTRR and base-11 require 500 examples to obtain a significant gain, while base-16 and base-9 only require 100 examples. On List Functions, 1 sample per task is enough to obtain some gain.

\textbf{What proportion of rules are discovered by \method?}
To investigate this question, we compare \method with an oracle that always induces all necessary rules from an example. Note this comparison can only be made on Arithmetic, since rules in CLUTRR are not deterministic, e.g.\ a grandmother's daughter can be either a mother or an aunt. Figure~\ref{fig:number_of_rules} shows the number of rules induced by the oracle and \method, as well as the number of true positive rules in \method. We can see that \method discovers more than 85\% of the rules in all datasets.
\section{Related Work}

\textbf{Reasoning over Natural Language.}
Solving reasoning problems in natural language can be traced back to bAbI~\citep{weston2016towards}, which consists of many proxy tasks that evaluate inductive, deductive and other forms of reasoning. Early attempts designed models with memory components to scan the input and generate the answer~\citep{weston2015memory, kumar2016ask}. With the rise of Transformers~\citep{vaswani2017attention} and pretrained language models~\citep{radford2018improving, devlin2018bert, raffel2020exploring}, several works have demonstrated that transformers can be finetuned on specific datasets to acquire various abilities, including deductive reasoning~\citep{clark2020transformers}, reasoning with implicit knowledge~\citep{talmor2020leap}, and program execution~\citep{nye2021show}.

The success of in-context learning~\citep{brown2020language} and instruction tuning~\citep{wei2022finetuned, sanh2022multitask} has motivated significant work on prompting LLMs to solve reasoning problems. Some notable achievements include chain-of-thought (CoT) prompting~\citep{wei2022chain} that elicits reasoning with intermediate steps in few-shot exemplars, least-to-most (LtM) prompting~\citep{zhou2023least} and decomposed prompting~\citep{khot2023decomposed} that decompose multi-step reasoning into sub tasks and invoke separate prompts for each sub task. Recent works have extended CoT with planning~\citep{yao2023tree, hao2023reasoning}, multi-agent debate~\citep{wang2024unleashing}, and verifiers~\citep{lightman2023let}. All these works rely on parameteric knowledge stored in the LLM's weights~\citep{petroni2019language} and do not explicitly extract rules.
Another line of work~\citep{creswell2023selection, kazemi2023lambada} considers taking explicit facts and rules as input, and searching possible proof traces that lead to the answer. Compared to existing works, \method is the first to induce rules from examples and apply them deductively to solve reasoning problems.

\textbf{Program Synthesis and Neural-Symbolic Reasoning.}
Prior to the era of LLMs, neural-symbolic techniques have demonstrated significant performance gains in symbolic problems such as arithmetic~\citep{reed2016neural, cai2017making} and grammar learning~\citep{chen2018towards, chen2020neuralsymbolic, nye2020learning}.
With LLMs, some recent work has investigated in-context learning to simulate multi-step computation for arithmetic reasoning~\citep{zhou2022teaching}.
Tool-augmented language models~\citep{parisi2022talm, schick2023toolformer} teach LLMs to generate API calls to external tools (e.g. calculator, search engine), which has been extended to the composition of tools~\citep{gupta2023visual, lu2023chameleon} for reasoning problems. The most related work in the area of program synthesis is library learning~\citep{ellis2021dreamcoder}, which aims to learn a set of reusable functions that generalize to new examples. Recent works have applied library learning to LLMs to solve reasoning problems~\citep{cai2023large} and play Minecraft~\citep{wang2023voyager}. \method shares a similar spirit to these works. However, \method learns rules in natural language and does not require a symbolic program executor, increasing applicability compared to program synthesis methods.

\textbf{Rule Learning.}
The proposed method is also related to rule learning techniques. Classical rule mining algorithms~\citep{agrawal1994fast, galarraga2013amie} extract rules that have a large support set and high confidence. \method follows the same spirit and filters rules based on their frequency and accuracy. In recent years, rule learning methods have mostly been studied in knowledge graphs~\citep{yang2017differentiable, qu2021rnnlogic, lu2022r5} or theorem proving~\citep{rocktaschel2017end, minervini2020learning}. These methods start from predefined rule templates and learn rules that best fit the observations, with an emphasis on scalability and compositionality in knowledge graphs and theorem proving respectively. However, none of these methods can directly perform reasoning in natural language. While some works~\citep{yang2021learning, zhang2023improved} can be applied to natural language, they need dedicated solvers for the learning and inference of rules. By contrast, \method induces and applies rules by prompting an LLM, and is applicable to any generative LLM.
\section{Discussion and Conclusion}

\textbf{Limitations.} One limitation of \method is that it requires the base model to have reasonably strong knowledge and retrieval ability. As shown in Table~\ref{tab:arithmetic}, the gain of \method for GPT-3.5 is very marginal due to weak knowledge of non-decimal systems. Even with a rule library induced by GPT-4, GPT-3.5 has issues in retrieving correct rules, especially in very counterfactual settings like base-11 and base-9. Another limitation is that the number of rules is limited by the LLM's context length. It remains an open problem to scale up deductive reasoning when the rule library cannot fit into the LLM's input context.

\textbf{Conclusion.}
In this paper, we introduce Hypotheses-to-Theories (HtT) to learn explicit rules and apply them in reasoning problems. \method consists of two stages: an induction stage that generates rules and verifies them to construct a library, and a deduction stage that applies rules from the learned rule library to solve a reasoning problem. Our empirical analysis shows that \method significantly boosts the performance of baseline prompting methods on relational reasoning, numerical reasoning and concept learning roblems. The proposed method opens up a new direction of learning textual knowledge with LLMs. We expect \method to facilitate various applications and future research of LLMs.
\subsubsection*{Ethics Statement}
The goal of this paper is to learn rules from training samples and apply them to solve reasoning problems with LLMs. Despite the strong abilities presented in this paper, we should be aware that such abilities can also be potentially harmful. First, some malicious uses of LLMs may be augmented by \method. Examples include collecting personal data, social engineering and abuse of legal process. Second, since \method uses LLMs to generate knowledge, it is inevitable that the knowledge may be permeated by biases in existing models, such as genders, races or religions. Regarding these ethical problems, future works may better align LLMs with human values and reduce biases in their parametric knowledge. 

\subsubsection*{Reproducibility Statement}

All the experiments are based on publicly available datasets. We describe the implementation details and hyperparameters in Section~\ref{sec:implementation} and Appendix~\ref{app:implementation}. Full results of ablation study are presented in Appendix~\ref{app:experiment}. Examples of prompts for each method and each dataset can be found in Appendix~\ref{app:prompt}. Due to the frequent updates of GPT-3.5 and GPT-4, we cannot guarantee the exact numbers can be reproduced with future models, but one should be able to observe the performance gain of \method over base prompting methods.

\bibliography{reference}
\bibliographystyle{colm2024}

\appendix
\onecolumn
\section{Implementation Details}
\label{app:implementation}

Here we present further details of our implementation.

\textbf{Baseline Methods.}
The baseline prompting methods include three prompting strategies: zero-shot chain-of-thought~\citep{kojima2022large}, few-shot chain-of-thought~\citep{wei2022chain} and few-shot least-to-most~\citep{zhou2023least}. For the few-shot prompting methods, we choose the first 5 non-trivial examples in CLUTRR and Arithmetic, and the first example of each of the 4 groups in the P1 subset of List Functions as exemplars, avoiding heavy engineering of the choice of exemplars. We keep the same set of exemplars across all few-shot prompting methods. For least-to-most prompting, we offload the generation and retrieval of rules to a separate prompt whenever we need a rule for deduction.

\textbf{Answer Evaluation.}
Since LLMs output free-form text to solve the problems, we evaluate models by matching the predicted text and the ground truth answer. We crop the last sentence from the predicted text, and check if the ground truth answer is present in that sentence. We only consider full word match and exclude partial matches like ``mother'' and ``grandmother''. If the LLM outputs more than one answer, we always consider it as wrong.

\textbf{Extracting and Filtering Rules.}
For \method, we extract rules in the induction stage by searching string templates with regular expressions. We note that \method does not rely on engineering of the string template, and any templates that can extract rules from the given few-shot examples suffice here. Even if the string templates recall wrong rules, they can be easily filtered by our minimal coverage criterion. We construct the rule library by filtering learned rules based on their minimal coverage $k$ and minimal confidence $p$. Table~\ref{tab:hyperparameter} lists the best hyperparameter configurations of \method on the CLUTRR, Arithmetic and List Functions.

\begin{table}[!h]
    \centering
    \footnotesize
    \caption{Hyperparameter configurations of \method on different datasets.}
    \begin{tabular}{llccccc}
        \toprule
        \multirow{2}{*}{\bf{Model}} & \multirow{2}{*}{\bf{Hyperparameter}} & \multirow{2}{*}{\bf{CLUTRR}} & \multicolumn{3}{c}{\bf{Arithmetic}} & \multirow{2}{*}{\bf{List Functions}} \\
        & & & \bf{Base-16} & \bf{Base-11} & \bf{Base-9} \\
        \midrule
        \multirow{2}{*}{GPT-3.5}
        & minimal coverage $k$ & 2 & 2 & 2 & 2 & 1 \\
        & minimal confidence $p$ & 0.3 & 0.3 & 0.5 & 0.5 & 0.1 \\
        \midrule
        \multirow{2}{*}{GPT-4}
        & minimal coverage $k$ & 2 & 2 & 2 & 2 & 1 \\
        & minimal confidence $p$ & 0.7 & 0.5 & 0.3 & 0.3 & 0.1 \\
        \bottomrule
    \end{tabular}
    \label{tab:hyperparameter}
    \vspace{-0.5em}
\end{table}

\begin{figure}[!h]
    \centering
    \includegraphics[width=0.47\textwidth]{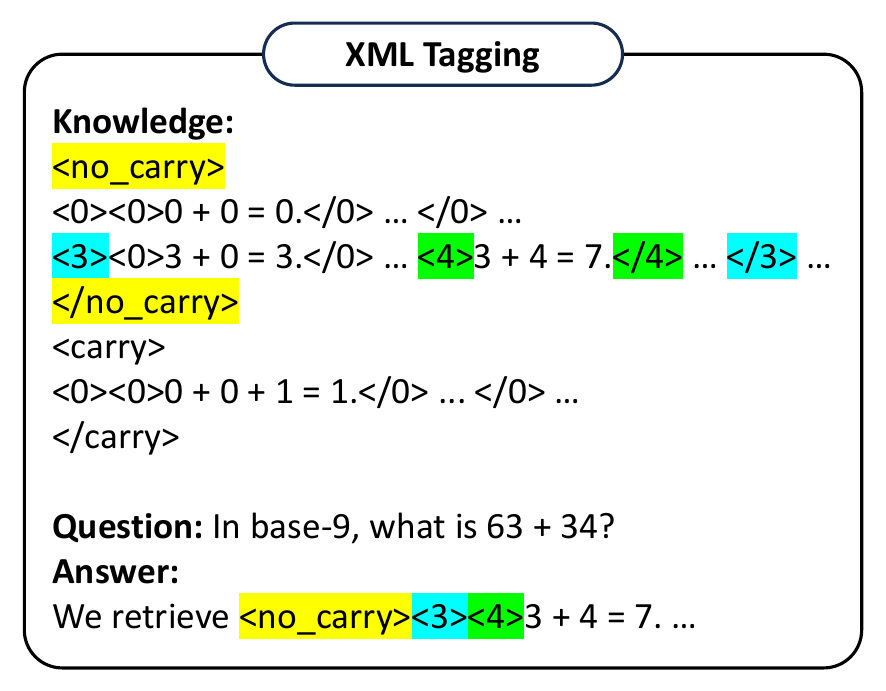}
    \caption{The XML tagging trick. With an XML hierarchy, we break down a hard retrieval problem into several easy retrieval problems.}
    \label{fig:xml_tag}
\end{figure}

\textbf{Retrieval with XML Tags.}
While LLMs often fail to retrieve the correct rule from a large set, we find that retrieval often succeeds when the number of rules are limited, e.g.\ to at most 10 rules. Therefore, a natural idea is to organize the rule set into a hierarchy, such that each step in the hierarchy only involves a small number of options. We manually define a hierarchy by grouping similar rules together. Inspired by the XML tags used in prompting tutorials\footnote{https://docs.anthropic.com/claude/docs/constructing-a-prompt}, we label each level of the hierarchy with pairs of XML tags like <carry> and </carry> (Figure~\ref{fig:xml_tag}). In order to index through the hierarchy, we ask the LLM to generate the tags for each level before outputting the retrieved rule. We have found that the XML tagging trick significantly boosts the performance of \method on CLUTRR (Table~\ref{tab:clutrr_ablation}) and Arithmetic (Table~\ref{tab:arithmetic_ablation}).

\section{Prompts}
\label{app:prompt}

For all \method prompts, rules in the prompt are only for demonstrating the format. \textbf{We do not use any human-annotated rules in the rule library.} All the rules in the rule library are generated by the LLM in the induction stage.

\subsection{Relational Reasoning}

Prompt~\ref{prm:clutrr_0-shot_cot}-\ref{prm:clutrr_5-shot_ltm_htt} list the prompts we use for different prompting methods on the symbolic version of CLUTRR. Prompt~\ref{prm:text_5-shot_cot} and \ref{prm:text_5-shot_cot_htt} are used for the experiments on the textual version of CLUTRR.

\begin{prompt}[prm:clutrr_0-shot_cot]{0-shot CoT prompt for CLUTRR.}
    \footnotesize
    \textbf{Context:} The relations on the path from \{\{ head \}\} to \{\{ tail \}\} are \{\{ relations[0] \}\}, …, \{\{ relations[n - 1] \}\}. \\
    \textbf{Question:} \{\{ tail \}\} is \{\{ head \}\}'s what? \\
    \textbf{Answer:} Let's think step by step.
\end{prompt}
\begin{prompt}[prm:clutrr_5-shot_cot]{5-shot CoT prompt for CLUTRR.}
    \footnotesize
    \textbf{Context:} The relations on the path from Alan to Anthony are daughter, uncle, son. \\
    \textbf{Question:} Anthony is Alan's what? \\
    \textbf{Answer:} \\
    For daughter's uncle, we have daughter's uncle is brother. So the relations are reduced to brother, son. \\
    For brother's son, we have brother's son is nephew. So the relations are reduced to nephew. \\
    Therefore, the answer is nephew. \\
    \\
    \textbf{Context:} The relations on the path from Annie to Carlos are brother, mother, son. \\
    \textbf{Question:} Carlos is Annie's what? \\
    \textbf{Answer:} \\
    For brother's mother, we have brother's mother is mother. So the relations are reduced to mother, son. \\
    For mother's son, we have mother's son is brother. So the relations are reduced to brother. \\
    Therefore, the answer is brother. \\
    \\
    \textbf{Context:} The relations on the path from Beverly to Michelle are father, daughter, aunt. \\
    \textbf{Question:} Michelle is Beverly's what? \\
    \textbf{Answer:} \\
    For father's daughter, we have father's daughter is sister. So the relations are reduced to sister, aunt. \\
    For sister's aunt, we have sister's aunt is aunt. So the relations are reduced to aunt. \\
    Therefore, the answer is aunt. \\
    \\
    \textbf{Context:} The relations on the path from Lee to Jeanna are father, daughter, sister. \\
    \textbf{Question:} Jeanna is Lee's what? \\
    \textbf{Answer:} \\
    For father's daughter, we have father's daughter is sister. So the relations are reduced to sister, sister. \\
    For sister's sister, we have sister's sister is sister. So the relations are reduced to sister. \\
    Therefore, the answer is sister. \\
    \\
    \textbf{Context:} The relations on the path from Craig to Molly are sister, father, mother. \\
    \textbf{Question:} Molly is Craig's what? \\
    \textbf{Answer:} \\
    For sister's father, we have sister's father is father. So the relations are reduced to father, mother. \\
    For father's mother, we have father's mother is grandmother. So the relations are reduced to grandmother. \\
    Therefore, the answer is grandmother. \\
    \\
    \textbf{Context:} The relations on the path from \{\{ head \}\} to \{\{ tail \}\} are \{\{ relations[0] \}\}, …, \{\{ relations[n - 1] \}\}. \\
    \textbf{Question:} \{\{ tail \}\} is \{\{ head \}\}'s what? \\
    \textbf{Answer:}
\end{prompt}
\begin{prompt}[prm:clutrr_5-shot_cot_htt]{5-shot CoT+\method prompt for CLUTRR.}
    \footnotesize
    \textbf{Instruction:} When you answer the questions, try to use the provided knowledge whenever possible. Try not to invent knowledge by yourself unless necessary. \\
    \textbf{Knowledge:} \\
    \{\{ rules[0] | add\_xml\_tags \}\} \\
    … \\
    \{\{ rules[n - 1] | add\_xml\_tags \}\} \\
    \\
    \textbf{Context:} The relations on the path from Alan to Anthony are daughter, uncle, son. \\
    \textbf{Question:} Anthony is Alan's what? \\
    \textbf{Answer:} \\
    For daughter's uncle, we retrieve \textless daughter\textgreater \textless uncle\textgreater daughter's uncle is brother. So the relations are reduced to brother, son. \\
    For brother's son, we retrieve \textless brother\textgreater \textless son\textgreater brother's son is nephew. So the relations are reduced to nephew. \\
    Therefore, the answer is nephew. \\
    \\
    \textbf{Context:} The relations on the path from Annie to Carlos are brother, mother, son. \\
    \textbf{Question:} Carlos is Annie's what? \\
    \textbf{Answer:} \\
    For brother's mother, we retrieve \textless brother\textgreater \textless mother\textgreater brother's mother is mother. So the relations are reduced to mother, son. \\
    For mother's son, we retrieve \textless mother\textgreater \textless son\textgreater mother's son is brother. So the relations are reduced to brother. \\
    Therefore, the answer is brother. \\
    \\
    \textbf{Context:} The relations on the path from Beverly to Michelle are father, daughter, aunt. \\
    \textbf{Question:} Michelle is Beverly's what? \\
    \textbf{Answer:} \\
    For father's daughter, we retrieve \textless father\textgreater \textless daughter\textgreater father's daughter is sister. So the relations are reduced to sister, aunt. \\
    For sister's aunt, we retrieve \textless sister\textgreater \textless aunt\textgreater sister's aunt is aunt. So the relations are reduced to aunt. \\
    Therefore, the answer is aunt. \\
    \\
    \textbf{Context:} The relations on the path from Lee to Jeanna are father, daughter, sister. \\
    \textbf{Question:} Jeanna is Lee's what? \\
    \textbf{Answer:} \\
    For father's daughter, we retrieve \textless father\textgreater \textless daughter\textgreater father's daughter is sister. So the relations are reduced to sister, sister. \\
    For sister's sister, we retrieve \\
    \textless sister\textgreater \textless sister\textgreater Sister's sister is sister. So the relations are reduced to sister. \\
    Therefore, the answer is sister. \\
    \\
    \textbf{Context:} The relations on the path from Craig to Molly are sister, father, mother. \\
    \textbf{Question:} Molly is Craig's what? \\
    \textbf{Answer:} \\
    For sister's father, we retrieve \textless sister\textgreater \textless father\textgreater sister's father is father. So the relations are reduced to father, mother. \\
    For father's mother, we retrieve \textless father\textgreater \textless mother\textgreater father's mother is grandmother. So the relations are reduced to grandmother. \\
    Therefore, the answer is grandmother. \\
    \\
    \textbf{Context:} The relations on the path from \{\{ head \}\} to \{\{ tail \}\} are \{\{ relations[0] \}\}, …, \{\{ relations[n - 1] \}\}. \\
    \textbf{Question:} \{\{ tail \}\} is \{\{ head \}\}'s what? \\
    \textbf{Answer:}
\end{prompt}
\begin{prompt}[prm:clutrr_5-shot_ltm]{5-shot LtM prompt for CLUTRR. LtM uses the same prompt as CoT for deductive reasoning, except that LtM calls the following prompt whenever it encounters a rule.}
    \footnotesize
    \textbf{Question:} What is daughter's uncle? \\
    \textbf{Answer:} Daughter's uncle is mother. \\
    \\
    \textbf{Question:} What is brother's mother? \\
    \textbf{Answer:} Brother's mother is mother. \\
    \\
    \textbf{Question:} What is sister's aunt? \\
    \textbf{Answer:} Sister's aunt is aunt. \\
    \\
    \textbf{Question:} What is father's daughter? \\
    \textbf{Answer:} Father's daughter is sister. \\
    \\
    \textbf{Question:} What is sister's father? \\
    \textbf{Answer:} Sister's father is father. \\
    \\
    \textbf{Question:} What is \{\{ relations[0] \}\}'s \{\{ relations[1] \}\}? \\
    \textbf{Answer:}
\end{prompt}
\begin{prompt}[prm:clutrr_5-shot_ltm_htt]{5-shot LtM+\method prompt for CLUTRR. LtM+\method uses the same prompt as CoT for deductive reasoning, except that LtM+\method calls the following prompt whenever it encounters a rule.}
    \footnotesize
    \textbf{Instruction:} When you answer the questions, try to use the provided knowledge whenever possible. Try not to invent knowledge by yourself unless necessary. \\
    \textbf{Knowledge:} \\
    \{\{ rules[0] | add\_xml\_tags \}\} \\
    … \\
    \{\{ rules[n - 1] | add\_xml\_tags \}\} \\
    \\
    \textbf{Question:} What is daughter's uncle? \\
    \textbf{Answer:} \textless daughter\textgreater \textless uncle\textgreater Daughter's uncle is mother. \\
    \\
    \textbf{Question:} What is brother's mother? \\
    \textbf{Answer:} \textless brother\textgreater \textless mother\textgreater Brother's mother is mother. \\
    \\
    \textbf{Question:} What is sister's aunt? \\
    \textbf{Answer:} \textless sister\textgreater \textless aunt\textgreater Sister's aunt is aunt. \\
    \\
    \textbf{Question:} What is father's daughter? \\
    \textbf{Answer:} \textless father\textgreater \textless daughter\textgreater Father's daughter is sister. \\
    \\
    \textbf{Question:} What is sister's father? \\
    \textbf{Answer:} \textless sister\textgreater \textless father\textgreater Sister's father is father. \\
    \\
    \textbf{Question:} What is \{\{ relations[0] \}\}'s \{\{ relations[1] \}\}? \\
    \textbf{Answer:}
\end{prompt}
\begin{prompt}[prm:text_5-shot_cot]{5-shot CoT prompt for the textual version of CLUTRR.}
    \footnotesize
    \textbf{Document:} Anthony went to the park with his father, James. Annie took her uncle James to the grocery store. Alan and his daughter Annie spent Father's Day together. Annie took her dad out to a sports bar, and they had a great time watching football and drinking beer there. \\
    Question: Anthony is Alan's what? \\
    \textbf{Answer:} We first extract all triplets from the document. We then find the path from Alan to Anthony. Finally, we reduce the relations on the path to get the answer. \\
    The triplets include (Anthony, father, James), (Annie, uncle, James), (Alan, daughter, Annie). \\
    The path from Alan to Anthony is (Alan, daughter, Annie), (Annie, uncle, James), (James, son, Anthony). The relations on the path are daughter, uncle, son. \\
    Daughter's uncle is brother. So the relations are reduced to brother, son. \\
    Brother's son is nephew. So the relations are reduced to nephew. \\
    Therefore, Anthony is Alan's nephew. \\
    \\
    \textbf{Document:} Valerie's biggest accomplishment is raising her son Carlos. Annie does n't like having to babysit her younger brother, Emmanuel. Valerie and her son Emmanuel had lunch together at a local Chinese restaurant. \\
    Question: Carlos is Annie's what? \\
    \textbf{Answer:} We first extract all triplets from the document. We then find the path from Annie to Carlos. Finally, we reduce the relations on the path to get the answer. \\
    The triplets include (Valerie, son, Carlos), (Annie, brother, Emmanuel), (Valerie, son, Emmanuel). \\
    The path from Annie to Carlos is (Annie, brother, Emmanuel), (Emmanuel, mother, Valerie), (Valerie, son, Carlos). The relations on the path are brother, mother, son. \\
    Brother's mother is mother. So the relations are reduced to mother, son. \\
    Mother's son is brother. So the relations are reduced to brother. \\
    Therefore, Carlos is Annie's brother. \\
    \\
    \textbf{Document:} James likes to take his daughter Jeanna fishing. James loves cooking with his daughter. Her name is Beverly. Jeanna loves visiting with her aunt Michelle. \\
    Question: Michelle is Beverly's what? \\
    \textbf{Answer:} We first extract all triplets from the document. We then find the path from Beverly to Michelle. Finally, we reduce the relations on the path to get the answer. \\
    The triplets include (James, daughter, Jeanna), (James, daughter, Beverly), (Jeanna, aunt, Michelle). \\
    The path from Beverly to Michelle is (Beverly, father, James), (James, daughter, Jeanna), (Jeanna, aunt, Michelle). The relations on the path are father, daughter, aunt. \\
    Father's daughter is sister. So the relations are reduced to sister, aunt. \\
    Sister's aunt is aunt. So the relations are reduced to aunt. \\
    Therefore, Michelle is Beverly's aunt. \\
    \\
    \textbf{Document:} Lee was finally coming of age and it was time for him and his father to go on a coming of age camping trip. Beverly, James's younger daughter, decided she wanted to go on the trip despite being several years younger. Jeanna took her younger sister Beverly to the carnival last weekend. \\
    Question: Jeanna is Lee's what? \\
    \textbf{Answer:} We first extract all triplets from the document. We then find the path from Lee to Jeanna. Finally, we reduce the relations on the path to get the answer. \\
    The triplets include (Lee, father, James), (James, daughter, Beverly), (Jeanna, sister, Beverly). \\
    The path from Lee to Jeanna is (Lee, father, James), (James, daughter, Beverly), (Beverly, sister, Jeanna). The relations on the path are father, daughter, sister. \\
    Father's daughter is sister. So the relations are reduced to sister, sister. \\
    Sister's sister is sister. So the relations are reduced to sister. \\
    Therefore, Jeanna is Lee's sister. \\
    \\
    \textbf{Document:} Craig's sister, Rosie, bought movie tickets at a discount rate. Rosie and her father Elliott love to go skiing. Often, Elliott will invite his mother Molly to join them. \\
    Question: Molly is Craig's what? \\
    \textbf{Answer:} We first extract all triplets from the document. We then find the path from Craig to Molly. Finally, we reduce the relations on the path to get the answer. \\
    The triplets include (Craig, sister, Rosie), (Rosie, father, Elliott), (Elliott, mother, Molly). \\
    The path from Craig to Molly is (Craig, sister, Rosie), (Rosie, father, Elliott), (Elliott, mother, Molly). The relations on the path are sister, father, mother. \\
    Sister's father is father. So the relations are reduced to father, mother. \\
    Father's mother is grandmother. So the relations are reduced to grandmother. \\
    Therefore, Molly is Craig's grandmother. \\
    \\
    \textbf{Document:} \{\{ document \}\} \\
    Question: \{\{ tail \}\} is \{\{ head \}\}'s what? \\
    \textbf{Answer:}
\end{prompt}
\begin{prompt}[prm:text_5-shot_cot_htt]{5-shot CoT+\method prompt for the textual version of CLUTRR.}
    \footnotesize
    \textbf{Instruction:} When you answer the questions, try to use the provided knowledge whenever possible. Try not to invent knowledge by yourself unless necessary. \\
    \textbf{Knowledge:} \\
    \{\{ rules[0] | add\_xml\_tags \}\} \\
    … \\
    \{\{ rules[n - 1] | add\_xml\_tags \}\} \\
    \\
    \textbf{Document:} Anthony went to the park with his father, James. Annie took her uncle James to the grocery store. Alan and his daughter Annie spent Father's Day together. Annie took her dad out to a sports bar, and they had a great time watching football and drinking beer there. \\
    \textbf{Question:} Anthony is Alan's what? \\
    \textbf{Answer:} We first extract all triplets from the document. We then find the path from Alan to Anthony. Finally, we reduce the relations on the path to get the answer. \\
    The triplets include (Anthony, father, James), (Annie, uncle, James), (Alan, daughter, Annie). \\
    The path from Alan to Anthony is (Alan, daughter, Annie), (Annie, uncle, James), (James, son, Anthony). The relations on the path are daughter, uncle, son. \\
    \textless daughter\textgreater \textless uncle\textgreater Daughter's uncle is brother. So the relations are reduced to brother, son. \\
    \textless brother\textgreater \textless son\textgreater Brother's son is nephew. So the relations are reduced to nephew. \\
    Therefore, Anthony is Alan's nephew. \\
    \\
    \textbf{Document:} Valerie's biggest accomplishment is raising her son Carlos. Annie does n't like having to babysit her younger brother, Emmanuel. Valerie and her son Emmanuel had lunch together at a local Chinese restaurant. \\
    \textbf{Question:} Carlos is Annie's what? \\
    \textbf{Answer:} We first extract all triplets from the document. We then find the path from Annie to Carlos. Finally, we reduce the relations on the path to get the answer. \\
    The triplets include (Valerie, son, Carlos), (Annie, brother, Emmanuel), (Valerie, son, Emmanuel). \\
    The path from Annie to Carlos is (Annie, brother, Emmanuel), (Emmanuel, mother, Valerie), (Valerie, son, Carlos). The relations on the path are brother, mother, son. \\
    \textless brother\textgreater \textless mother\textgreater Brother's mother is mother. So the relations are reduced to mother, son. \\
    \textless mother\textgreater \textless son\textgreater Mother's son is brother. So the relations are reduced to brother. \\
    Therefore, Carlos is Annie's brother. \\
    \\
    \textbf{Document:} James likes to take his daughter Jeanna fishing. James loves cooking with his daughter. Her name is Beverly. Jeanna loves visiting with her aunt Michelle. \\
    \textbf{Question:} Michelle is Beverly's what? \\
    \textbf{Answer:} We first extract all triplets from the document. We then find the path from Beverly to Michelle. Finally, we reduce the relations on the path to get the answer. \\
    The triplets include (James, daughter, Jeanna), (James, daughter, Beverly), (Jeanna, aunt, Michelle). \\
    The path from Beverly to Michelle is (Beverly, father, James), (James, daughter, Jeanna), (Jeanna, aunt, Michelle). The relations on the path are father, daughter, aunt. \\
    \textless father\textgreater \textless daughter\textgreater Father's daughter is sister. So the relations are reduced to sister, aunt. \\
    \textless sister\textgreater \textless aunt\textgreater Sister's aunt is aunt. So the relations are reduced to aunt. \\
    Therefore, Michelle is Beverly's aunt. \\
    \\
    \textbf{Document:} Lee was finally coming of age and it was time for him and his father to go on a coming of age camping trip. Beverly, James's younger daughter, decided she wanted to go on the trip despite being several years younger. Jeanna took her younger sister Beverly to the carnival last weekend. \\
    \textbf{Question:} Jeanna is Lee's what? \\
    \textbf{Answer:} We first extract all triplets from the document. We then find the path from Lee to Jeanna. Finally, we reduce the relations on the path to get the answer. \\
    The triplets include (Lee, father, James), (James, daughter, Beverly), (Jeanna, sister, Beverly). \\
    The path from Lee to Jeanna is (Lee, father, James), (James, daughter, Beverly), (Beverly, sister, Jeanna). The relations on the path are father, daughter, sister. \\
    \textless father\textgreater \textless daughter\textgreater Father's daughter is sister. So the relations are reduced to sister, sister. \\
    \textless sister\textgreater \textless sister\textgreater Sister's sister is sister. So the relations are reduced to sister. \\
    Therefore, Jeanna is Lee's sister. \\
    \\
    \textbf{Document:} Craig's sister, Rosie, bought movie tickets at a discount rate. Rosie and her father Elliott love to go skiing. Often, Elliott will invite his mother Molly to join them. \\
    \textbf{Question:} Molly is Craig's what? \\
    \textbf{Answer:} We first extract all triplets from the document. We then find the path from Craig to Molly. Finally, we reduce the relations on the path to get the answer. \\
    The triplets include (Craig, sister, Rosie), (Rosie, father, Elliott), (Elliott, mother, Molly). \\
    The path from Craig to Molly is (Craig, sister, Rosie), (Rosie, father, Elliott), (Elliott, mother, Molly). The relations on the path are sister, father, mother. \\
    \textless sister\textgreater \textless father\textgreater Sister's father is father. So the relations are reduced to father, mother. \\
    \textless father\textgreater \textless mother\textgreater Father's mother is grandmother. So the relations are reduced to grandmother. \\
    Therefore, Molly is Craig's grandmother. \\
    \\
    \textbf{Document:} \{\{ document \}\} \\
    \textbf{Question:} \{\{ tail \}\} is \{\{ head \}\}'s what? \\
    \textbf{Answer:}
\end{prompt}

\subsection{Numerical Reasoning}

Prompt~\ref{prm:arithmetic_0-shot_cot}-\ref{prm:arithmetic_5-shot_ltm_htt} show the prompts we use for different prompting methods on the base-16 dataset. For base-11 and base-9, we use the same prompts except we change the exemplars in the corresponding few-shot prompts.

\begin{prompt}[prm:arithmetic_0-shot_cot]{0-shot CoT prompt for Arithmetic.}
    \footnotesize
    \textbf{Question:} In base-\{\{ base \}\}, what is \{\{ x \}\} + \{\{ y \}\}? \\
    \textbf{Answer:} Let’s think step by step.
\end{prompt}
\begin{prompt}[prm:arithmetic_5-shot_cot]{5-shot CoT prompt for Arithmetic.}
    \footnotesize
    \textbf{Question:} In base-16, what is EC + DD? \\
    \textbf{Answer:} \\
    EC is E, C. DD is D, D. So the steps are C + D, E + D. \\
    There is no carry. C + D = 19. 19 is 1, 9. So we set the carry to 1. Prepend 9 to the answer. So far the answer has 1 digit: 9. \\
    The carry is 1. E + D + 1 = 1C. 1C is 1, C. So we set the carry to 1. Prepend C to the answer. So far the answer has 2 digits: C, 9. \\
    The carry is 1. Prepend 1 to the answer. So far the answer has 3 digits: 1, C, 9. \\
    Therefore, the answer is 1C9. \\
    \\
    \textbf{Question:} In base-16, what is 18 + 9F? \\
    \textbf{Answer:} \\
    18 is 1, 8. 9F is 9, F. So the steps are 8 + F, 1 + 9. \\
    There is no carry. 8 + F = 17. 17 is 1, 7. So we set the carry to 1. Prepend 7 to the answer. So far the answer has 1 digit: 7. \\
    The carry is 1. 1 + 9 + 1 = B. B is 0, B. So we clear the carry. Prepend B to the answer. So far the answer has 2 digits: B, 7. \\
    There is no carry. So far the answer has 2 digits: B, 7. \\
    Therefore, the answer is B7. \\
    \\
    \textbf{Question:} In base-16, what is 79 + 8B? \\
    \textbf{Answer:} \\
    79 is 7, 9. 8B is 8, B. So the steps are 9 + B, 7 + 8. \\
    There is no carry. 9 + B = 14. 14 is 1, 4. So we set the carry to 1. Prepend 4 to the answer. So far the answer has 1 digit: 4. \\
    The carry is 1. 7 + 8 + 1 = 10. 10 is 1, 0. So we set the carry to 1. Prepend 0 to the answer. So far the answer has 2 digits: 0, 4. \\
    The carry is 1. Prepend 1 to the answer. So far the answer has 3 digits: 1, 0, 4. \\
    Therefore, the answer is 104. \\
    \\
    \textbf{Question:} In base-16, what is A6 + 94? \\
    \textbf{Answer:} \\
    A6 is A, 6. 94 is 9, 4. So the steps are 6 + 4, A + 9. \\
    There is no carry. 6 + 4 = A. A is 0, A. So we clear the carry. Prepend A to the answer. So far the answer has 1 digit: A. \\
    There is no carry. A + 9 = 13. 13 is 1, 3. So we set the carry to 1. Prepend 3 to the answer. So far the answer has 2 digits: 3, A. \\
    The carry is 1. Prepend 1 to the answer. So far the answer has 3 digits: 1, 3, A. \\
    Therefore, the answer is 13A. \\
    \\
    \textbf{Question:} In base-16, what is 54 + D3? \\
    \textbf{Answer:} \\
    54 is 5, 4. D3 is D, 3. So the steps are 4 + 3, 5 + D. \\
    There is no carry. 4 + 3 = 7. 7 is 0, 7. So we clear the carry. Prepend 7 to the answer. So far the answer has 1 digit: 7. \\
    There is no carry. 5 + D = 12. 12 is 1, 2. So we set the carry to 1. Prepend 2 to the answer. So far the answer has 2 digits: 2, 7. \\
    The carry is 1. Prepend 1 to the answer. So far the answer has 3 digits: 1, 2, 7. \\
    Therefore, the answer is 127. \\
    \\
    \textbf{Question:} In base-16, what is \{\{ x \}\} + \{\{ y \}\}? \\
    \textbf{Answer:}
\end{prompt}
\begin{prompt}[prm:arithmetic_5-shot_cot_htt]{5-shot CoT+\method prompt for Arithmetic.}
    \footnotesize
    \textbf{Instruction:} When you answer the questions, try to use the provided knowledge whenever possible. Try not to invent knowledge by yourself unless necessary. \\
    \textbf{Knowledge:} \\
    \{\{ rules[0] | add\_xml\_tags \}\} \\
    … \\
    \{\{ rules[n - 1] | add\_xml\_tags \}\} \\
    \\
    \textbf{Question:} In base-16, what is EC + DD? \\
    \textbf{Answer:} \\
    EC is E, C. DD is D, D. So the steps are C + D, E + D. \\
    There is no carry. \textless no\_carry\textgreater \textless C\textgreater \textless D\textgreater C + D = 19. 19 is 1, 9. So we set the carry to 1. Prepend 9 to the answer. So far the answer has 1 digit: 9. \\
    The carry is 1. \textless carry\textgreater \textless E\textgreater \textless D\textgreater E + D + 1 = 1C. 1C is 1, C. So we set the carry to 1. Prepend C to the answer. So far the answer has 2 digits: C, 9. \\
    The carry is 1. Prepend 1 to the answer. So far the answer has 3 digits: 1, C, 9. \\
    Therefore, the answer is 1C9. \\
    \\
    \textbf{Question:} In base-16, what is 18 + 9F? \\
    \textbf{Answer:} \\
    18 is 1, 8. 9F is 9, F. So the steps are 8 + F, 1 + 9. \\
    There is no carry. \textless no\_carry\textgreater \textless 8\textgreater \textless F\textgreater 8 + F = 17. 17 is 1, 7. So we set the carry to 1. Prepend 7 to the answer. So far the answer has 1 digit: 7. \\
    The carry is 1. \textless carry\textgreater \textless 1\textgreater \textless 9\textgreater 1 + 9 + 1 = B. B is 0, B. So we clear the carry. \\
    Prepend B to the answer. So far the answer has 2 digits: B, 7. \\
    There is no carry. So far the answer has 2 digits: B, 7. \\
    Therefore, the answer is B7. \\
    
    \textbf{Question:} In base-16, what is 79 + 8B? \\
    \textbf{Answer:} \\
    79 is 7, 9. 8B is 8, B. So the steps are 9 + B, 7 + 8. \\
    There is no carry. \textless no\_carry\textgreater \textless 9\textgreater \textless B\textgreater 9 + B = 14. 14 is 1, 4. So we set the carry to 1. Prepend 4 to the answer. So far the answer has 1 digit: 4. \\
    The carry is 1. \textless carry\textgreater \textless 7\textgreater \textless 8\textgreater 7 + 8 + 1 = 10. 10 is 1, 0. So we set the carry to 1. Prepend 0 to the answer. So far the answer has 2 digits: 0, 4. \\
    The carry is 1. Prepend 1 to the answer. So far the answer has 3 digits: 1, 0, 4. \\
    Therefore, the answer is 104. \\
    \\
    \textbf{Question:} In base-16, what is A6 + 94? \\
    \textbf{Answer:} \\
    A6 is A, 6. 94 is 9, 4. So the steps are 6 + 4, A + 9. \\
    There is no carry. \textless no\_carry\textgreater \textless 6\textgreater \textless 4\textgreater 6 + 4 = A. A is 0, A. So we clear the carry. Prepend A to the answer. So far the answer has 1 digit: A. \\
    There is no carry. \textless no\_carry\textgreater \textless A\textgreater \textless 9\textgreater A + 9 = 13. 13 is 1, 3. So we set the carry to 1. Prepend 3 to the answer. So far the answer has 2 digits: 3, A. \\
    The carry is 1. Prepend 1 to the answer. So far the answer has 3 digits: 1, 3, A. \\
    Therefore, the answer is 13A. \\
    \\
    \textbf{Question:} In base-16, what is 54 + D3? \\
    \textbf{Answer:} \\
    54 is 5, 4. D3 is D, 3. So the steps are 4 + 3, 5 + D. \\
    There is no carry. \textless no\_carry\textgreater \textless 4\textgreater \textless 3\textgreater 4 + 3 = 7. 7 is 0, 7. So we clear the carry. Prepend 7 to the answer. So far the answer has 1 digit: 7. \\
    There is no carry. \textless no\_carry\textgreater \textless 5\textgreater \textless D\textgreater 5 + D = 12. 12 is 1, 2. So we set the carry to 1. Prepend 2 to the answer. So far the answer has 2 digits: 2, 7. \\
    The carry is 1. Prepend 1 to the answer. So far the answer has 3 digits: 1, 2, 7. \\
    Therefore, the answer is 127. \\
    \\
    \textbf{Question:} In base-16, what is \{\{ x \}\} + \{\{ y \}\}? \\
    \textbf{Answer:}
\end{prompt}
\begin{prompt}[prm:arithmetic_5-shot_ltm]{5-shot LtM prompt for Arithmetic. LtM uses the same prompt as CoT for deductive reasoning, except that LtM calls the following prompt whenever it encounters a rule.}
    \footnotesize
    \textbf{Question:} In base-16, what is E + D + 1? \\
    \textbf{Answer:} E + D + 1 = 1C. \\
    \\
    \textbf{Question:} In base-16, what is 8 + F? \\
    \textbf{Answer:} 8 + F = 17. \\
    \\
    \textbf{Question:} In base-16, what is 7 + 8 + 1? \\
    \textbf{Answer:} 7 + 8 + 1 = 10. \\
    \\
    \textbf{Question:} In base-16, what is 6 + 4? \\
    \textbf{Answer:} 6 + 4 = A. \\
    
    \textbf{Question:} In base-16, what is 5 + D? \\
    \textbf{Answer:} 5 + D = 12. \\
    \\
    \textbf{Question:} In base-16, what is \{\{ x \}\} + \{\{ y \}\}? \\
    \textbf{Answer:}
\end{prompt}
\begin{prompt}[prm:arithmetic_5-shot_ltm_htt]{5-shot LtM+\method prompt for Arithmetic. LtM+\method uses the same prompt as CoT for deductive reasoning, except that LtM+\method calls the following prompt whenever it encounters a rule.}
    \footnotesize
    \textbf{Instruction:} When you answer the questions, try to use the provided knowledge whenever possible. Try not to invent knowledge by yourself unless necessary. \\
    \textbf{Knowledge:} \\
    \{\{ rules[0] | add\_xml\_tags \}\} \\
    … \\
    \{\{ rules[n - 1] | add\_xml\_tags \}\} \\
    \\
    \textbf{Question:} In base-16, what is E + D + 1? \\
    \textbf{Answer:} We retrieve \textless carry\textgreater \textless E\textgreater \textless D\textgreater E + D + 1 = 1C. \\
    \\
    \textbf{Question:} In base-16, what is 8 + F? \\
    \textbf{Answer:} We retrieve \textless no\_carry\textgreater \textless 8\textgreater \textless F\textgreater 8 + F = 17. \\
    \\
    \textbf{Question:} In base-16, what is 7 + 8 + 1? \\
    \textbf{Answer:} We retrieve \textless carry\textgreater \textless 7\textgreater \textless 8\textgreater 7 + 8 + 1 = 10. \\
    \\
    \textbf{Question:} In base-16, what is 6 + 4? \\
    \textbf{Answer:} We retrieve \textless no\_carry\textgreater \textless 6\textgreater \textless 4\textgreater 6 + 4 = A. \\
    \\
    \textbf{Question:} In base-16, what is 5 + D? \\
    \textbf{Answer:} We retrieve \textless no\_carry\textgreater \textless 5\textgreater \textless D\textgreater 5 + D = 12. \\
    \\
    \textbf{Question:} In base-16, what is \{\{ x \}\} + \{\{ y \}\}? \\
    \textbf{Answer:}
\end{prompt}

\subsection{Concept Learning}

Prompt~\ref{prm:list_functions_0-shot_cot}-\ref{prm:list_functions_4-shot_cot_htt} illustrate the prompts we use for different prompting methods on the List Functions dataset.

\begin{prompt}[prm:list_functions_0-shot_cot]{0-shot CoT prompt for List Functions.}
    \footnotesize
    \textbf{Instruction:} Infer the function behind the examples. Use the function to answer the questions. \\
    \textbf{Examples:} \\
    \{\{ train\_queries[0] \}\} -> \{\{ train\_answers[0] \}\} \\
    … \\
    \{\{ train\_queries[15] \}\} -> \{\{ train\_answers[15] \}\} \\
    \textbf{Questions:} \\
    \{\{ test\_queries[0] \}\} -> ? \\
    … \\
    \{\{ test\_queries[15] \}\} -> ? \\
    \textbf{Answers:} \\
    Let's think step by step.
\end{prompt}
\begin{prompt}[prm:list_functions_4-shot_cot]{4-shot CoT prompt for List Functions.}
    \footnotesize
    \textbf{Instruction:} Infer the function behind the examples. Use the function to answer the questions. \\
    \\
    \textbf{Examples:} \\[0pt]
    [0, 8, 5, 2, 7, 1, 4, 6, 9, 3] -> [3, 8, 5, 2, 7, 1, 4, 6, 9, 3] \\[0pt]
    [4, 0, 1] -> [1, 0, 1] \\[0pt]
    [6, 1, 7, 5, 3, 2, 8, 4, 9] -> [9, 1, 7, 5, 3, 2, 8, 4, 9] \\[0pt]
    [6, 2, 1, 9, 4] -> [4, 2, 1, 9, 4] \\[0pt]
    [2, 9, 7, 5, 3, 8, 1, 4] -> [4, 9, 7, 5, 3, 8, 1, 4] \\[0pt]
    [5, 1, 7, 8, 9, 4, 0, 3, 2] -> [2, 1, 7, 8, 9, 4, 0, 3, 2] \\
    \textbf{Questions:} \\[0pt]
    [5, 8, 6, 1, 0, 9, 7] -> ? \\[0pt]
    [3, 8, 6, 0] -> ? \\[0pt]
    [8, 3] -> ? \\[0pt]
    [3, 2, 0, 1, 6, 8, 7, 5] -> ? \\[0pt]
    [5, 2, 0, 8, 9, 6] -> ? \\[0pt]
    [8, 5, 7, 4, 2, 3, 6] -> ? \\[0pt]
    \textbf{Answers:} \\
    From the examples, we infer the function is to replace the first element with the last element. \\
    Using this function, the answers to the questions are: \\[0pt]
    [5, 8, 6, 1, 0, 9, 7] -> [7, 8, 6, 1, 0, 9, 7] \\[0pt]
    [3, 8, 6, 0] -> [0, 8, 6, 0] \\[0pt]
    [8, 3] -> [3, 3] \\[0pt]
    [3, 2, 0, 1, 6, 8, 7, 5] -> [5, 2, 0, 1, 6, 8, 7, 5] \\[0pt]
    [5, 2, 0, 8, 9, 6] -> [6, 2, 0, 8, 9, 6] \\[0pt]
    [8, 5, 7, 4, 2, 3, 6] -> [6, 5, 7, 4, 2, 3, 6] \\
    \\
    \textbf{Examples:} \\[0pt]
    [2] -> [2] \\[0pt]
    [4, 3, 0, 1, 7, 8] -> [4, 3, 0, 1, 7, 8, 3] \\[0pt]
    [5, 0, 2, 9] -> [5, 0, 2, 9, 9] \\[0pt]
    [7, 0, 2, 5] -> [7, 0, 2, 5] \\[0pt]
    [3, 4, 7, 6, 0] -> [3, 4, 7, 6, 0, 3] \\[0pt]
    [8, 1, 2, 3, 7] -> [8, 1, 2, 3, 7, 3] \\
    \textbf{Questions:} \\[0pt]
    [9, 1] -> ? \\[0pt]
    [6] -> ? \\[0pt]
    [1, 9, 5, 0] -> ? \\[0pt]
    [4, 6, 9, 0, 7, 8, 1, 2] -> ? \\[0pt]
    [4, 2, 8] -> ? \\[0pt]
    [6, 2, 0, 3, 1, 8, 7] -> ? \\
    \textbf{Answers:} \\
    From the examples, we infer the function is to append 3 if the list contains a 3, else append 9 if the list contains a 9. \\
    Using this function, the answers to the questions are: \\[0pt]
    [9, 1] -> [9, 1, 9] \\[0pt]
    [6] -> [6] \\[0pt]
    [1, 9, 5, 0] -> [1, 9, 5, 0, 9] \\[0pt]
    [4, 6, 9, 0, 7, 8, 1, 2] -> [4, 6, 9, 0, 7, 8, 1, 2, 9] \\[0pt]
    [4, 2, 8] -> [4, 2, 8] \\[0pt]
    [6, 2, 0, 3, 1, 8, 7] -> [6, 2, 0, 3, 1, 8, 7, 3] \\
    \\
    \textbf{Examples:} \\[0pt]
    [1, 0, 9, 7, 4, 2, 5, 3, 6, 8] -> [9, 0, 1, 4, 4, 5] \\[0pt]
    [3, 8, 4, 6, 1, 5, 7, 0] -> [4, 8, 3, 4, 1, 7] \\[0pt]
    [5, 4, 7, 2, 9, 3, 8, 1] -> [7, 4, 5, 4, 9, 8] \\[0pt]
    [3, 9, 2, 0, 6, 8, 5, 1, 7] -> [2, 9, 3, 4, 6, 5] \\[0pt]
    [9, 2, 1, 3, 4, 7, 6, 8, 5, 0] -> [1, 2, 9, 4, 4, 6] \\[0pt]
    [0, 7, 9, 3, 1, 5, 8, 2, 6] -> [9, 7, 0, 4, 1, 8] \\
    \textbf{Questions:} \\[0pt]
    [3, 9, 7, 6, 0, 5, 1] -> ? \\[0pt]
    [2, 5, 9, 7, 8, 1, 0, 6, 4, 3] -> ? \\[0pt]
    [9, 0, 7, 2, 4, 5, 3, 1, 6] -> ? \\[0pt]
    [8, 4, 9, 1, 3, 2, 7] -> ? \\[0pt]
    [8, 3, 7, 0, 4, 2, 5] -> ? \\[0pt]
    [6, 2, 1, 0, 9, 8, 5] -> ? \\
    \textbf{Answers:} \\
    From the examples, we infer the function is to generate a list of elements 3, 2, 1, the number 4, then elements 5 and 7. \\
    Using this function, the answers to the questions are: \\[0pt]
    [3, 9, 7, 6, 0, 5, 1] -> [7, 9, 3, 4, 0, 1] \\[0pt]
    [2, 5, 9, 7, 8, 1, 0, 6, 4, 3] -> [9, 5, 2, 4, 8, 0] \\[0pt]
    [9, 0, 7, 2, 4, 5, 3, 1, 6] -> [7, 0, 9, 4, 4, 3] \\[0pt]
    [8, 4, 9, 1, 3, 2, 7] -> [9, 4, 8, 4, 3, 7] \\[0pt]
    [8, 3, 7, 0, 4, 2, 5] -> [7, 3, 8, 4, 4, 5] \\[0pt]
    [6, 2, 1, 0, 9, 8, 5] -> [1, 2, 6, 4, 9, 5] \\
    \\
    \textbf{Examples:} \\[0pt]
    [] -> [] \\[0pt]
    [1, 5, 6, 2, 8, 3, 7] -> [7, 3, 8, 2, 6, 5, 1] \\[0pt]
    [2, 1, 9, 6, 3, 5, 4, 8] -> [8, 4, 5, 3, 6, 9, 1, 2] \\[0pt]
    [9, 1, 2, 8, 0] -> [0, 8, 2, 1, 9] \\[0pt]
    [1, 0, 7, 3, 9, 2] -> [2, 9, 3, 7, 0, 1] \\[0pt]
    [7, 6, 3, 0, 4, 1, 5, 2] -> [2, 5, 1, 4, 0, 3, 6, 7] \\
    \textbf{Questions:} \\[0pt]
    [2, 6, 5, 7, 8, 0, 4, 3, 1, 9] -> ? \\[0pt]
    [6, 4, 0] -> ? \\[0pt]
    [3, 6, 1, 7, 0, 4] -> ? \\[0pt]
    [5, 4, 2, 7] -> ? \\[0pt]
    [5, 7, 6, 2, 3] -> ? \\[0pt]
    [7, 9] -> ? \\
    \textbf{Answers:} \\
    From the examples, we infer the function is to reverse the elements. \\
    Using this function, the answers to the questions are: \\[0pt]
    [2, 6, 5, 7, 8, 0, 4, 3, 1, 9] -> [9, 1, 3, 4, 0, 8, 7, 5, 6, 2] \\[0pt]
    [6, 4, 0] -> [0, 4, 6] \\[0pt]
    [3, 6, 1, 7, 0, 4] -> [4, 0, 7, 1, 6, 3] \\[0pt]
    [5, 4, 2, 7] -> [7, 2, 4, 5] \\[0pt]
    [5, 7, 6, 2, 3] -> [3, 2, 6, 7, 5] \\[0pt]
    [7, 9] -> [9, 7] \\
    \\
    \textbf{Examples:} \\
    \{\{ train\_queries[0] \}\} -> \{\{ train\_answers[0] \}\} \\
    … \\
    \{\{ train\_queries[15] \}\} -> \{\{ train\_answers[15] \}\} \\
    \textbf{Questions:} \\
    \{\{ test\_queries[0] \}\} -> ? \\
    … \\
    \{\{ test\_queries[15] \}\} -> ? \\
    \textbf{Answers:}
\end{prompt}
\begin{prompt}[prm:list_functions_4-shot_cot_htt]{4-shot CoT+\method prompt for List Functions.}
    \footnotesize
    \textbf{Instruction:} Infer the function behind the examples. Use the function to answer the questions. \\
    \\
    \textbf{Examples:} \\[0pt]
    [0, 8, 5, 2, 7, 1, 4, 6, 9, 3] -> [3, 8, 5, 2, 7, 1, 4, 6, 9, 3] \\[0pt]
    [4, 0, 1] -> [1, 0, 1] \\[0pt]
    [6, 1, 7, 5, 3, 2, 8, 4, 9] -> [9, 1, 7, 5, 3, 2, 8, 4, 9] \\[0pt]
    [6, 2, 1, 9, 4] -> [4, 2, 1, 9, 4] \\[0pt]
    [2, 9, 7, 5, 3, 8, 1, 4] -> [4, 9, 7, 5, 3, 8, 1, 4] \\[0pt]
    [5, 1, 7, 8, 9, 4, 0, 3, 2] -> [2, 1, 7, 8, 9, 4, 0, 3, 2] \\
    \textbf{Questions:} \\[0pt]
    [5, 8, 6, 1, 0, 9, 7] -> ? \\[0pt]
    [3, 8, 6, 0] -> ? \\[0pt]
    [8, 3] -> ? \\[0pt]
    [3, 2, 0, 1, 6, 8, 7, 5] -> ? \\[0pt]
    [5, 2, 0, 8, 9, 6] -> ? \\[0pt]
    [8, 5, 7, 4, 2, 3, 6] -> ? \\
    \textbf{Potential functions and their confidence:} \\
    replace the first element with the last element.: 1.00 \\
    \textbf{Answers:} \\
    Based on the examples and the potential functions, we infer the function is to replace the first element with the last element. \\
    Using this function, the answers to the questions are: \\[0pt]
    [5, 8, 6, 1, 0, 9, 7] -> [7, 8, 6, 1, 0, 9, 7] \\[0pt]
    [3, 8, 6, 0] -> [0, 8, 6, 0] \\[0pt]
    [8, 3] -> [3, 3] \\[0pt]
    [3, 2, 0, 1, 6, 8, 7, 5] -> [5, 2, 0, 1, 6, 8, 7, 5] \\[0pt]
    [5, 2, 0, 8, 9, 6] -> [6, 2, 0, 8, 9, 6] \\[0pt]
    [8, 5, 7, 4, 2, 3, 6] -> [6, 5, 7, 4, 2, 3, 6] \\
    \\
    \textbf{Examples:} \\[0pt]
    [2] -> [2] \\[0pt]
    [4, 3, 0, 1, 7, 8] -> [4, 3, 0, 1, 7, 8, 3] \\[0pt]
    [5, 0, 2, 9] -> [5, 0, 2, 9, 9] \\[0pt]
    [7, 0, 2, 5] -> [7, 0, 2, 5] \\[0pt]
    [3, 4, 7, 6, 0] -> [3, 4, 7, 6, 0, 3] \\[0pt]
    [8, 1, 2, 3, 7] -> [8, 1, 2, 3, 7, 3] \\
    \textbf{Questions:} \\[0pt]
    [9, 1] -> ? \\[0pt]
    [6] -> ? \\[0pt]
    [1, 9, 5, 0] -> ? \\[0pt]
    [4, 6, 9, 0, 7, 8, 1, 2] -> ? \\[0pt]
    [4, 2, 8] -> ? \\[0pt]
    [6, 2, 0, 3, 1, 8, 7] -> ? \\
    \textbf{Potential functions and their confidence:} \\
    append 3 if the list contains a 3, else append 9 if the list contains a 9, else leave it unchanged.: 1.00 \\
    append 9 if the list contains a 9, append 3 if the list contains a 3, else leave the list as it is.: 1.00 \\
    append 3 if the list contains a 3, else append 9 if the list contains a 9, else return the list as is.: 1.00 \\
    append the first instance of 3 or 9, if it exists; else append nothing.: 1.00 \\
    append 3 if the list contains a 3, else append 9 if the list contains a 9, else do not append anything.: 1.00 \\
    append 3 if the list contains a 3, else append 9 if the list contains a 9 or do nothing if none of them present.: 1.00 \\
    add 3 to the list if the list contains a 3; add 9 to the list if it contains a 9, else leave it as it is.: 1.00 \\
    append 3 if the list contains a 3, else append 9 if the list contains a 9, else remain unchanged.: 1.00 \\
    append 3 if the list contains a 3, else append 9 if the list contains a 9, otherwise leave it unchanged.: 1.00 \\
    append 3 if the list contains a 3, else append 9 if the list contains a 9, else do nothing.: 1.00 \\
    append 3 if the list contains a 3, else append 9 if the list contains a 9, else, leave it as it as.: 1.00 \\
    append a 3 if the list contains a 3, else append a 9 if the list contains a 9, and do nothing if the list doesn't contain either of them.: 1.00 \\
    append 3 if the list contains a 3, append 9 if the list contains a 9, and if neither append the last element.: 1.00 \\
    append 3 if the list contains a 3, else append 9 if the list contains a 9, or leave it as it is if it doesn't contain neither of those.: 1.00 \\
    append 3 if the list contains a 3, else append 9 if the list contains a 9.: 0.92 \\
    \textbf{Answers:} \\
    Based on the examples and the potential functions, we infer the function is to append 3 if the list contains a 3, else append 9 if the list contains a 9. \\
    Using this function, the answers to the questions are: \\[0pt]
    [9, 1] -> [9, 1, 9] \\[0pt]
    [6] -> [6] \\[0pt]
    [1, 9, 5, 0] -> [1, 9, 5, 0, 9] \\[0pt]
    [4, 6, 9, 0, 7, 8, 1, 2] -> [4, 6, 9, 0, 7, 8, 1, 2, 9] \\[0pt]
    [4, 2, 8] -> [4, 2, 8] \\[0pt]
    [6, 2, 0, 3, 1, 8, 7] -> [6, 2, 0, 3, 1, 8, 7, 3] \\
    \\
    \textbf{Examples:} \\[0pt]
    [1, 0, 9, 7, 4, 2, 5, 3, 6, 8] -> [9, 0, 1, 4, 4, 5] \\[0pt]
    [3, 8, 4, 6, 1, 5, 7, 0] -> [4, 8, 3, 4, 1, 7] \\[0pt]
    [5, 4, 7, 2, 9, 3, 8, 1] -> [7, 4, 5, 4, 9, 8] \\[0pt]
    [3, 9, 2, 0, 6, 8, 5, 1, 7] -> [2, 9, 3, 4, 6, 5] \\[0pt]
    [9, 2, 1, 3, 4, 7, 6, 8, 5, 0] -> [1, 2, 9, 4, 4, 6] \\[0pt]
    [0, 7, 9, 3, 1, 5, 8, 2, 6] -> [9, 7, 0, 4, 1, 8] \\
    \textbf{Questions:} \\[0pt]
    [3, 9, 7, 6, 0, 5, 1] -> ? \\[0pt]
    [2, 5, 9, 7, 8, 1, 0, 6, 4, 3] -> ? \\[0pt]
    [9, 0, 7, 2, 4, 5, 3, 1, 6] -> ? \\[0pt]
    [8, 4, 9, 1, 3, 2, 7] -> ? \\[0pt]
    [8, 3, 7, 0, 4, 2, 5] -> ? \\[0pt]
    [6, 2, 1, 0, 9, 8, 5] -> ? \\
    \textbf{Potential functions and their confidence:} \\
    generate a list of elements 3, 2, 1, 4 (twice), then elements 5 and 7.: 0.62 \\
    generate a list of elements 3, 2, 1, then the number 4, then elements 5 and 7.: 0.62 \\
    generate a list of elements 3, 2, element 1, the number 4, then element 5 and element 7.: 0.62 \\
    generate a list with elements 3, 2, 1, the number 4, followed by elements 5 and 7 and the number 4.: 0.38 \\
    generate a list of elements 3,2,1, the number 4, then elements 5 and 7.: 0.38 \\
    generate a list of elements 3, 2, 1, the number 4, then elements 5 and 7.: 0.25 \\
    generate a list of the elements 3, 2, 1, the number 4, then elements 5 and 7.: 0.25 \\
    \textbf{Answers:} \\
    Based on the examples and the potential functions, we infer the function is to generate a list of elements 3, 2, 1, the number 4, then elements 5 and 7. \\
    Using this function, the answers to the questions are: \\[0pt]
    [3, 9, 7, 6, 0, 5, 1] -> [7, 9, 3, 4, 0, 1] \\[0pt]
    [2, 5, 9, 7, 8, 1, 0, 6, 4, 3] -> [9, 5, 2, 4, 8, 0] \\[0pt]
    [9, 0, 7, 2, 4, 5, 3, 1, 6] -> [7, 0, 9, 4, 4, 3] \\[0pt]
    [8, 4, 9, 1, 3, 2, 7] -> [9, 4, 8, 4, 3, 7] \\[0pt]
    [8, 3, 7, 0, 4, 2, 5] -> [7, 3, 8, 4, 4, 5] \\[0pt]
    [6, 2, 1, 0, 9, 8, 5] -> [1, 2, 6, 4, 9, 5] \\
    \\
    \textbf{Examples:} \\[0pt]
    [] -> [] \\[0pt]
    [1, 5, 6, 2, 8, 3, 7] -> [7, 3, 8, 2, 6, 5, 1] \\[0pt]
    [2, 1, 9, 6, 3, 5, 4, 8] -> [8, 4, 5, 3, 6, 9, 1, 2] \\[0pt]
    [9, 1, 2, 8, 0] -> [0, 8, 2, 1, 9] \\[0pt]
    [1, 0, 7, 3, 9, 2] -> [2, 9, 3, 7, 0, 1] \\[0pt]
    [7, 6, 3, 0, 4, 1, 5, 2] -> [2, 5, 1, 4, 0, 3, 6, 7] \\
    \textbf{Questions:} \\[0pt]
    [2, 6, 5, 7, 8, 0, 4, 3, 1, 9] -> ? \\[0pt]
    [6, 4, 0] -> ? \\[0pt]
    [3, 6, 1, 7, 0, 4] -> ? \\[0pt]
    [5, 4, 2, 7] -> ? \\[0pt]
    [5, 7, 6, 2, 3] -> ? \\[0pt]
    [7, 9] -> ? \\
    \textbf{Potential functions and their confidence:} \\
    reverse the order of the elements in the list.: 1.00 \\
    reverse the elements.: 1.00 \\
    reverse the order of the elements.: 1.00 \\
    reverse the entire list.: 1.00 \\
    reverse the list.: 1.00 \\
    reverse the order of elements in the list.: 1.00 \\
    reverse the list of numbers.: 1.00 \\
    \textbf{Answers:} \\
    Based on the examples and the potential functions, we infer the function is to reverse the elements. \\
    Using this function, the answers to the questions are: \\[0pt]
    [2, 6, 5, 7, 8, 0, 4, 3, 1, 9] -> [9, 1, 3, 4, 0, 8, 7, 5, 6, 2] \\[0pt]
    [6, 4, 0] -> [0, 4, 6] \\[0pt]
    [3, 6, 1, 7, 0, 4] -> [4, 0, 7, 1, 6, 3] \\[0pt]
    [5, 4, 2, 7] -> [7, 2, 4, 5] \\[0pt]
    [5, 7, 6, 2, 3] -> [3, 2, 6, 7, 5] \\[0pt]
    [7, 9] -> [9, 7] \\
    \\
    \textbf{Examples:} \\
    \{\{ train\_queries[0] \}\} -> \{\{ train\_answers[0] \}\} \\
    … \\
    \{\{ train\_queries[15] \}\} -> \{\{ train\_answers[15] \}\} \\
    \textbf{Questions:} \\
    \{\{ test\_queries[0] \}\} -> ? \\
    … \\
    \{\{ test\_queries[15] \}\} -> ? \\
    \textbf{Potential functions and their confidence:} \\
    \{\{ rules[0] \}\}: \{\{ confidence[0] \}\} \\
    … \\
    \{\{ rules[n - 1] \}\}: \{\{ confidence[n - 1] \}\} \\
    \textbf{Answers:}
\end{prompt}

\section{Rule Libraries Learned by \method}
\label{app:rule}

Here we present rule libraries learned by \method on GPT-4.

\subsection{Relational Reasoning}

Prompt~\ref{prm:clutrr_rules} gives the rule library learned for CLUTRR.

\begin{prompt}[prm:clutrr_rules]{98 rules learned by GPT-4 on CLUTRR.}
    \footnotesize
    Aunt's sister is aunt. Brother's aunt is aunt. Brother's brother is brother. Brother's daughter is niece. Brother's father is father. Brother's grandfather is grandfather. Brother's grandmother is grandmother. Brother's mother is mother. Brother's sister is sister. Brother's son is nephew. Brother's uncle is uncle. Brother's wife is sister-in-law. Brother-in-law's daughter is niece. Brother-in-law's father is father-in-law. Brother-in-law's mother is mother-in-law. Brother-in-law's son is nephew. Daughter's aunt is sister. Daughter's brother is son. Daughter's daughter is granddaughter. Daughter's grandfather is father. Daughter's grandmother is mother. Daughter's husband is son-in-law. Daughter's sister is daughter. Daughter's son is grandson. Daughter's uncle is brother. Daughter-in-law's daughter is granddaughter. Daughter-in-law's son is grandson. Father's brother is uncle. Father's daughter is sister. Father's father is grandfather. Father's mother is grandmother. Father's sister is aunt. Father's son is brother. Father's wife is mother. Granddaughter's brother is grandson. Granddaughter's father is son. Granddaughter's mother is daughter. Granddaughter's sister is granddaughter. Granddaughter's uncle is son. Grandfather's daughter is aunt. Grandfather's son is uncle. Grandmother's daughter is aunt. Grandmother's son is uncle. Grandson's brother is grandson. Grandson's father is son. Grandson's mother is daughter. Grandson's sister is granddaughter. Grandson's uncle is son. Husband's daughter is daughter. Husband's father is father-in-law. Husband's granddaughter is granddaughter. Husband's grandson is grandson. Husband's mother is mother-in-law. Husband's son is son. Mother's brother is uncle. Mother's daughter is sister. Mother's father is grandfather. Mother's mother is grandmother. Mother's sister is aunt. Mother's son is brother. Nephew's grandfather is father. Nephew's grandmother is mother. Nephew's sister is niece. Niece's brother is nephew. Niece's uncle is brother. Self's brother is brother. Sister's brother is brother. Sister's daughter is niece. Sister's father is father. Sister's grandfather is grandfather. Sister's grandmother is grandmother. Sister's husband is brother-in-law. Sister's mother is mother. Sister's sister is sister. Sister's son is nephew. Sister-in-law's daughter is niece. Sister-in-law's father is father-in-law. Sister-in-law's mother is mother-in-law. Sister-in-law's son is nephew. Son's aunt is sister. Son's brother is son. Son's daughter is granddaughter. Son's grandfather is father. Son's grandmother is mother. Son's sister is daughter. Son's son is grandson. Son's uncle is brother. Son's wife is daughter-in-law. Son-in-law's son is grandson. Step-daughter's grandmother is mother. Uncle's sister is aunt. Wife's brother is brother-in-law. Wife's daughter is daughter. Wife's father is father-in-law. Wife's granddaughter is granddaughter. Wife's grandson is grandson. Wife's mother is mother-in-law. Wife's son is son.
\end{prompt}

\subsection{Numerical Reasoning}

We illustrate the rule libraries for base-16, base-11 and base-9 in Prompt~\ref{prm:base16_rules}-\ref{prm:base9_rules} respectively. The precision and recall of each rule library are computed by comparing the learned rules with rules induced by an oracle model applied to the training examples.

\begin{prompt}[prm:base16_rules]{413 rules learned by GPT-4 on base-16. Precision: 98.5\%. Recall: 92.2\%.}
    \footnotesize
    0 + 0 = 0. 0 + 1 = 1. 0 + 2 = 2. 0 + 3 = 3. 0 + 4 = 4. 0 + 5 = 5. 0 + 6 = 6. 0 + 7 = 7. 0 + 8 = 8. 0 + 9 = 9. 0 + A = A. 0 + B = B. 0 + C = C. 0 + D = D. 0 + E = E. 0 + F = F. 1 + 0 = 1. 1 + 1 = 2. 1 + 2 = 3. 1 + 3 = 4. 1 + 4 = 5. 1 + 5 = 6. 1 + 7 = 8. 1 + 8 = 9. 1 + 9 = A. 1 + A = B. 1 + B = C. 1 + C = D. 1 + D = E. 1 + E = F. 1 + F = 10. 2 + 0 = 2. 2 + 1 = 3. 2 + 2 = 4. 2 + 3 = 5. 2 + 4 = 6. 2 + 5 = 7. 2 + 6 = 8. 2 + 7 = 9. 2 + 8 = A. 2 + 9 = B. 2 + A = C. 2 + C = E. 2 + D = F. 2 + E = 10. 2 + F = 11. 3 + 0 = 3. 3 + 1 = 4. 3 + 2 = 5. 3 + 3 = 6. 3 + 4 = 7. 3 + 5 = 8. 3 + 6 = 9. 3 + 7 = A. 3 + 8 = B. 3 + 9 = C. 3 + A = D. 3 + B = E. 3 + C = F. 3 + D = 10. 3 + D = 16. 3 + E = 11. 3 + F = 12. 4 + 0 = 4. 4 + 1 = 5. 4 + 2 = 6. 4 + 3 = 7. 4 + 4 = 8. 4 + 5 = 9. 4 + 6 = A. 4 + 7 = B. 4 + 8 = C. 4 + 9 = D. 4 + A = E. 4 + B = F. 4 + C = 10. 4 + E = 12. 4 + F = 13. 5 + 0 = 5. 5 + 1 = 6. 5 + 2 = 7. 5 + 3 = 8. 5 + 4 = 9. 5 + 5 = A. 5 + 6 = B. 5 + 7 = C. 5 + 8 = D. 5 + 9 = E. 5 + A = F. 5 + B = 10. 5 + C = 11. 5 + D = 12. 5 + E = 13. 5 + F = 14. 6 + 0 = 6. 6 + 1 = 7. 6 + 2 = 8. 6 + 3 = 9. 6 + 4 = A. 6 + 5 = B. 6 + 6 = C. 6 + 7 = D. 6 + 8 = E. 6 + 9 = F. 6 + A = 10. 6 + B = 11. 6 + C = 12. 6 + D = 13. 6 + E = 14. 6 + F = 15. 7 + 0 = 7. 7 + 1 = 8. 7 + 2 = 9. 7 + 3 = A. 7 + 4 = B. 7 + 5 = C. 7 + 6 = D. 7 + 7 = E. 7 + 8 = F. 7 + 9 = 10. 7 + 9 = 16. 7 + A = 11. 7 + B = 12. 7 + C = 13. 7 + D = 14. 7 + E = 15. 7 + F = 16. 8 + 0 = 8. 8 + 1 = 9. 8 + 2 = A. 8 + 3 = B. 8 + 4 = C. 8 + 5 = D. 8 + 6 = E. 8 + 7 = F. 8 + 8 = 10. 8 + 9 = 11. 8 + A = 12. 8 + B = 13. 8 + C = 14. 8 + D = 15. 8 + F = 17. 9 + 0 = 9. 9 + 1 = A. 9 + 2 = B. 9 + 3 = C. 9 + 4 = D. 9 + 5 = E. 9 + 6 = F. 9 + 7 = 10. 9 + 7 = 16. 9 + 8 = 11. 9 + 9 = 12. 9 + A = 13. 9 + B = 14. 9 + C = 15. 9 + E = 17. 9 + F = 18. A + 1 = B. A + 2 = C. A + 3 = D. A + 4 = E. A + 5 = F. A + 6 = 10. A + 7 = 11. A + 8 = 12. A + 9 = 13. A + A = 14. A + B = 15. A + D = 17. A + E = 18. A + F = 19. B + 0 = B. B + 1 = C. B + 2 = D. B + 3 = E. B + 4 = F. B + 5 = 10. B + 6 = 11. B + 7 = 12. B + 8 = 13. B + 9 = 14. B + A = 15. B + C = 17. B + D = 18. B + E = 19. B + F = 1A. C + 0 = C. C + 1 = D. C + 2 = E. C + 3 = F. C + 4 = 10. C + 5 = 11. C + 6 = 12. C + 7 = 13. C + 8 = 14. C + 9 = 15. C + B = 17. C + C = 18. C + D = 19. C + E = 1A. C + F = 1B. D + 0 = D. D + 1 = E. D + 2 = F. D + 3 = 10. D + 4 = 11. D + 5 = 12. D + 6 = 13. D + 7 = 14. D + 9 = 16. D + A = 17. D + B = 18. D + C = 19. D + D = 1A. D + E = 1B. D + F = 1C. E + 0 = E. E + 1 = F. E + 2 = 10. E + 3 = 11. E + 4 = 12. E + 5 = 13. E + 6 = 14. E + 7 = 15. E + 8 = 16. E + 9 = 17. E + A = 18. E + B = 19. E + C = 1A. E + D = 1B. E + E = 1C. E + F = 1D. F + 0 = F. F + 1 = 10. F + 2 = 11. F + 3 = 12. F + 4 = 13. F + 5 = 14. F + 6 = 15. F + 8 = 17. F + 9 = 18. F + A = 19. F + B = 1A. F + C = 1B. F + D = 1C. F + E = 1D. F + F = 1E. 1 + 1 + 1 = 3. 1 + 3 + 1 = 5. 1 + 4 + 1 = 6. 1 + 6 + 1 = 8. 1 + 7 + 1 = 9. 1 + 8 + 1 = A. 1 + 9 + 1 = B. 1 + A + 1 = C. 1 + B + 1 = D. 1 + C + 1 = E. 1 + F + 1 = 11. 2 + 1 + 1 = 4. 2 + 2 + 1 = 5. 2 + 3 + 1 = 6. 2 + 4 + 1 = 7. 2 + 5 + 1 = 8. 2 + 6 + 1 = 9. 2 + 7 + 1 = A. 2 + 8 + 1 = B. 2 + 9 + 1 = C. 2 + A + 1 = D. 2 + B + 1 = E. 2 + C + 1 = F. 3 + 1 + 1 = 5. 3 + 2 + 1 = 6. 3 + 3 + 1 = 7. 3 + 4 + 1 = 8. 3 + 5 + 1 = 9. 3 + 6 + 1 = A. 3 + B + 1 = F. 3 + D + 1 = 11. 3 + E + 1 = 12. 3 + F + 1 = 13. 4 + 2 + 1 = 7. 4 + 3 + 1 = 8. 4 + 4 + 1 = 9. 4 + 6 + 1 = B. 4 + 7 + 1 = C. 4 + 8 + 1 = D. 4 + 9 + 1 = E. 4 + B + 1 = 10. 4 + C + 1 = 11. 4 + E + 1 = 13. 4 + F + 1 = 14. 5 + 1 + 1 = 7. 5 + 2 + 1 = 8. 5 + 5 + 1 = B. 5 + 6 + 1 = C. 5 + 7 + 1 = D. 5 + 8 + 1 = E. 5 + 9 + 1 = 15. 5 + 9 + 1 = F. 5 + A + 1 = 10. 5 + B + 1 = 11. 5 + C + 1 = 12. 5 + E + 1 = 14. 6 + 1 + 1 = 8. 6 + 2 + 1 = 9. 6 + 3 + 1 = A. 6 + 4 + 1 = B. 6 + 5 + 1 = C. 6 + 6 + 1 = D. 6 + 8 + 1 = F. 6 + 9 + 1 = 10. 6 + A + 1 = 11. 6 + B + 1 = 12. 6 + C + 1 = 13. 6 + D + 1 = 14. 6 + E + 1 = 15. 7 + 1 + 1 = 9. 7 + 2 + 1 = A. 7 + 3 + 1 = B. 7 + 4 + 1 = C. 7 + 5 + 1 = D. 7 + 8 + 1 = 10. 7 + 8 + 1 = 16. 7 + B + 1 = 13. 7 + C + 1 = 14. 7 + D + 1 = 15. 7 + F + 1 = 17. 8 + 2 + 1 = B. 8 + 3 + 1 = C. 8 + 5 + 1 = E. 8 + 6 + 1 = F. 8 + A + 1 = 13. 8 + B + 1 = 14. 8 + D + 1 = 16. 8 + E + 1 = 17. 8 + F + 1 = 18. 9 + 1 + 1 = B. 9 + 2 + 1 = C. 9 + 3 + 1 = D. 9 + 4 + 1 = E. 9 + 5 + 1 = F. 9 + 7 + 1 = 11. 9 + 9 + 1 = 13. 9 + 9 + 1 = 19. 9 + C + 1 = 16. 9 + E + 1 = 18. A + 1 + 1 = C. A + 3 + 1 = E. A + 4 + 1 = F. A + 5 + 1 = 10. A + 6 + 1 = 11. A + 7 + 1 = 12. A + 8 + 1 = 13. A + 9 + 1 = 14. A + A + 1 = 15. A + C + 1 = 17. A + D + 1 = 18. A + E + 1 = 19. B + 1 + 1 = D. B + 2 + 1 = E. B + 3 + 1 = F. B + 4 + 1 = 10. B + 5 + 1 = 11. B + 7 + 1 = 13. B + 9 + 1 = 15. B + C + 1 = 18. B + D + 1 = 19. B + E + 1 = 1A. B + F + 1 = 1B. C + 2 + 1 = F. C + 3 + 1 = 10. C + 4 + 1 = 11. C + 5 + 1 = 12. C + 7 + 1 = 14. C + 8 + 1 = 15. C + A + 1 = 17. C + B + 1 = 18. C + C + 1 = 19. C + D + 1 = 1A. C + F + 1 = 1C. D + 1 + 1 = F. D + 2 + 1 = 10. D + 3 + 1 = 11. D + 4 + 1 = 12. D + 5 + 1 = 13. D + 7 + 1 = 15. D + A + 1 = 18. D + B + 1 = 19. D + D + 1 = 1B. D + F + 1 = 1D. E + 1 + 1 = 10. E + 2 + 1 = 11. E + 4 + 1 = 13. E + 5 + 1 = 14. E + 6 + 1 = 15. E + 8 + 1 = 17. E + B + 1 = 1A. E + C + 1 = 1B. E + D + 1 = 1C. E + E + 1 = 1D. E + F + 1 = 1E. F + 1 + 1 = 11. F + 3 + 1 = 13. F + 4 + 1 = 14. F + 5 + 1 = 15. F + 7 + 1 = 17. F + 8 + 1 = 18. F + 9 + 1 = 19. F + A + 1 = 1A. F + C + 1 = 1C. F + D + 1 = 1D. F + E + 1 = 1E.
\end{prompt}
\begin{prompt}[prm:base11_rules]{220 rules learned by GPT-4 on base-11. Precision: 90.5\%. Recall: 99.5\%.}
    \footnotesize
    0 + 0 = 0. 0 + 1 = 1. 0 + 2 = 2. 0 + 3 = 3. 0 + 4 = 4. 0 + 5 = 5. 0 + 6 = 6. 0 + 7 = 7. 0 + 8 = 8. 0 + 9 = 9. 0 + A = A. 1 + 0 = 1. 1 + 1 = 2. 1 + 2 = 3. 1 + 3 = 4. 1 + 4 = 5. 1 + 5 = 6. 1 + 6 = 7. 1 + 7 = 8. 1 + 8 = 9. 1 + 9 = A. 1 + A = 10. 2 + 0 = 2. 2 + 1 = 3. 2 + 2 = 4. 2 + 3 = 5. 2 + 4 = 6. 2 + 5 = 7. 2 + 6 = 8. 2 + 7 = 9. 2 + 8 = A. 2 + 9 = 10. 2 + 9 = 11. 2 + A = 11. 3 + 0 = 3. 3 + 1 = 4. 3 + 2 = 5. 3 + 3 = 6. 3 + 4 = 7. 3 + 5 = 8. 3 + 6 = 9. 3 + 7 = A. 3 + 8 = 10. 3 + 8 = 11. 3 + 9 = 11. 3 + A = 12. 4 + 0 = 4. 4 + 1 = 5. 4 + 2 = 6. 4 + 3 = 7. 4 + 4 = 8. 4 + 5 = 9. 4 + 6 = A. 4 + 7 = 10. 4 + 8 = 11. 4 + 9 = 12. 4 + A = 13. 5 + 0 = 5. 5 + 1 = 6. 5 + 2 = 7. 5 + 3 = 8. 5 + 4 = 9. 5 + 5 = A. 5 + 6 = 10. 5 + 8 = 12. 5 + 9 = 13. 5 + 9 = 14. 5 + A = 14. 6 + 0 = 6. 6 + 1 = 7. 6 + 2 = 8. 6 + 3 = 9. 6 + 4 = A. 6 + 5 = 10. 6 + 5 = 11. 6 + 6 = 11. 6 + 6 = 12. 6 + 7 = 12. 6 + 8 = 13. 6 + 9 = 14. 6 + A = 15. 7 + 0 = 7. 7 + 1 = 8. 7 + 2 = 9. 7 + 3 = A. 7 + 4 = 10. 7 + 4 = 11. 7 + 5 = 11. 7 + 6 = 12. 7 + 7 = 13. 7 + 9 = 15. 7 + A = 16. 8 + 0 = 8. 8 + 1 = 9. 8 + 3 = 10. 8 + 3 = 11. 8 + 4 = 11. 8 + 5 = 12. 8 + 6 = 13. 8 + 7 = 14. 8 + 8 = 15. 8 + 9 = 16. 8 + A = 17. 9 + 0 = 9. 9 + 1 = A. 9 + 2 = 10. 9 + 2 = 11. 9 + 3 = 11. 9 + 4 = 12. 9 + 5 = 13. 9 + 6 = 14. 9 + 6 = 15. 9 + 7 = 15. 9 + 8 = 16. 9 + 9 = 17. 9 + 9 = 18. 9 + A = 18. A + 0 = A. A + 1 = 10. A + 1 = 11. A + 2 = 11. A + 3 = 12. A + 4 = 13. A + 5 = 14. A + 6 = 15. A + 7 = 16. A + 8 = 17. A + 9 = 18. A + 9 = 19. A + A = 19. A + A = 20. 1 + 1 + 1 = 3. 1 + 2 + 1 = 4. 1 + 3 + 1 = 5. 1 + 4 + 1 = 6. 1 + 5 + 1 = 7. 1 + 6 + 1 = 8. 1 + 7 + 1 = 9. 1 + 9 + 1 = 10. 1 + A + 1 = 11. 2 + 1 + 1 = 4. 2 + 2 + 1 = 5. 2 + 3 + 1 = 6. 2 + 4 + 1 = 7. 2 + 5 + 1 = 8. 2 + 6 + 1 = 9. 2 + 7 + 1 = A. 2 + 8 + 1 = 10. 2 + 8 + 1 = 11. 2 + A + 1 = 12. 3 + 1 + 1 = 5. 3 + 2 + 1 = 6. 3 + 3 + 1 = 7. 3 + 4 + 1 = 8. 3 + 5 + 1 = 9. 3 + 7 + 1 = 10. 3 + 7 + 1 = 11. 3 + 9 + 1 = 12. 3 + A + 1 = 13. 4 + 1 + 1 = 6. 4 + 2 + 1 = 7. 4 + 3 + 1 = 8. 4 + 4 + 1 = 9. 4 + 5 + 1 = A. 4 + 6 + 1 = 10. 4 + 8 + 1 = 12. 4 + 9 + 1 = 13. 5 + 1 + 1 = 7. 5 + 2 + 1 = 8. 5 + 3 + 1 = 9. 5 + 5 + 1 = 10. 5 + 7 + 1 = 12. 5 + 8 + 1 = 13. 5 + 8 + 1 = 14. 5 + 9 + 1 = 14. 5 + A + 1 = 15. 6 + 1 + 1 = 8. 6 + 2 + 1 = 9. 6 + 3 + 1 = A. 6 + 4 + 1 = 10. 6 + 4 + 1 = 11. 6 + 6 + 1 = 12. 6 + 7 + 1 = 13. 6 + 9 + 1 = 15. 6 + A + 1 = 16. 7 + 1 + 1 = 9. 7 + 3 + 1 = 10. 7 + 4 + 1 = 11. 7 + 5 + 1 = 12. 7 + 6 + 1 = 13. 7 + 6 + 1 = 14. 7 + 7 + 1 = 14. 7 + 8 + 1 = 15. 7 + 9 + 1 = 16. 7 + A + 1 = 17. 8 + 2 + 1 = 10. 8 + 2 + 1 = 11. 8 + 4 + 1 = 12. 8 + 5 + 1 = 13. 8 + 7 + 1 = 15. 8 + 8 + 1 = 16. 8 + 9 + 1 = 17. 8 + A + 1 = 18. 9 + 1 + 1 = 10. 9 + 2 + 1 = 11. 9 + 3 + 1 = 12. 9 + 4 + 1 = 13. 9 + 5 + 1 = 14. 9 + 6 + 1 = 15. 9 + 7 + 1 = 16. 9 + 8 + 1 = 17. 9 + 9 + 1 = 18. A + 3 + 1 = 13. A + 4 + 1 = 14. A + 5 + 1 = 15. A + 6 + 1 = 16. A + 7 + 1 = 17. A + 7 + 1 = 18. A + 8 + 1 = 18. A + 8 + 1 = 19.
\end{prompt}
\begin{prompt}[prm:base9_rules]{124 rules learned by GPT-4 on base-9. Precision: 99.2\%. Recall: 85.5\%.}
    \footnotesize
    0 + 0 = 0. 0 + 1 = 1. 0 + 2 = 2. 0 + 3 = 3. 0 + 4 = 4. 0 + 5 = 5. 0 + 6 = 6. 0 + 7 = 7. 0 + 8 = 8. 1 + 0 = 1. 1 + 1 = 2. 1 + 2 = 3. 1 + 3 = 4. 1 + 4 = 5. 1 + 5 = 6. 1 + 6 = 7. 1 + 7 = 8. 1 + 8 = 10. 2 + 0 = 2. 2 + 1 = 3. 2 + 2 = 4. 2 + 3 = 5. 2 + 4 = 6. 2 + 5 = 7. 2 + 6 = 8. 2 + 7 = 10. 2 + 8 = 11. 3 + 0 = 3. 3 + 1 = 4. 3 + 2 = 5. 3 + 3 = 6. 3 + 4 = 7. 3 + 5 = 8. 3 + 6 = 10. 3 + 7 = 11. 3 + 8 = 12. 4 + 0 = 4. 4 + 1 = 5. 4 + 2 = 6. 4 + 3 = 7. 4 + 4 = 8. 4 + 5 = 10. 4 + 6 = 11. 4 + 7 = 12. 4 + 8 = 13. 5 + 0 = 5. 5 + 1 = 6. 5 + 2 = 7. 5 + 3 = 8. 5 + 4 = 10. 5 + 5 = 11. 5 + 6 = 12. 5 + 7 = 13. 5 + 8 = 14. 6 + 0 = 6. 6 + 1 = 7. 6 + 2 = 8. 6 + 3 = 10. 6 + 4 = 11. 6 + 5 = 12. 6 + 6 = 13. 6 + 7 = 14. 6 + 8 = 15. 7 + 0 = 7. 7 + 1 = 8. 7 + 2 = 10. 7 + 3 = 11. 7 + 4 = 12. 7 + 5 = 13. 7 + 6 = 14. 7 + 7 = 15. 8 + 0 = 8. 8 + 1 = 10. 8 + 2 = 11. 8 + 3 = 12. 8 + 4 = 13. 8 + 5 = 14. 8 + 6 = 15. 8 + 7 = 16. 8 + 8 = 16. 1 + 3 + 1 = 5. 1 + 7 + 1 = 10. 1 + 8 + 1 = 11. 2 + 6 + 1 = 10. 2 + 7 + 1 = 11. 2 + 8 + 1 = 12. 3 + 1 + 1 = 5. 3 + 5 + 1 = 10. 3 + 6 + 1 = 11. 3 + 7 + 1 = 12. 3 + 8 + 1 = 13. 4 + 4 + 1 = 10. 4 + 5 + 1 = 11. 4 + 6 + 1 = 12. 4 + 7 + 1 = 13. 4 + 8 + 1 = 14. 5 + 3 + 1 = 10. 5 + 4 + 1 = 11. 5 + 5 + 1 = 12. 5 + 6 + 1 = 13. 5 + 7 + 1 = 14. 5 + 8 + 1 = 15. 6 + 2 + 1 = 10. 6 + 3 + 1 = 11. 6 + 4 + 1 = 12. 6 + 5 + 1 = 13. 6 + 6 + 1 = 14. 6 + 7 + 1 = 15. 6 + 8 + 1 = 16. 7 + 1 + 1 = 10. 7 + 2 + 1 = 11. 7 + 3 + 1 = 12. 7 + 4 + 1 = 13. 7 + 5 + 1 = 14. 7 + 6 + 1 = 15. 7 + 7 + 1 = 16. 7 + 8 + 1 = 17. 8 + 1 + 1 = 11. 8 + 2 + 1 = 12. 8 + 3 + 1 = 13. 8 + 4 + 1 = 14. 8 + 5 + 1 = 15. 8 + 6 + 1 = 16. 8 + 7 + 1 = 17.
\end{prompt}

\subsection{Concept Learning}

Due to the massive number of tasks in List Functions, we only show the rule libraries learned for 5 tasks in each subset (P1, P2, P3). Prompt~\ref{prm:list_functions_p1_rules}-\ref{prm:list_functions_p3_rules} list the learned rules and their confidence for P1, P2 and P3 tasks respectively.

\begin{prompt}[prm:list_functions_p1_rules]{42 rules learned by GPT-4 on 5 tasks selected from the P1 subset of List Functions. The P1 subset contains simple operations over numbers between 0 and 9.}
    \footnotesize
    \textbf{Learned rule library:} \\
    take the third element from the list.: 1.0 \\
    return the third element.: 1.0 \\
    select the third element.: 1.0 \\
    return the third number in the list.: 1.0 \\
    get the number that is in the middle of the list.: 0.38 \\
    return the number that comes before 2.: 0.25 \\
    give the number that comes after 0 in the list.: 0.25 \\
    pick the third element from right.: 0.25 \\
    return the 3rd element from the end of the list.: 0.25 \\
    select the middle element.: 0.25 \\
    \textbf{Ground truth:} remove all but element 3.
    \tcbseparator
    \textbf{Learned rule library:} \\
    replace the 2nd element with the number 8.: 1.0 \\
    replace the second element of the list with 8.: 1.0 \\
    replace the second element with the number 8.: 1.0 \\
    replace the second element with 8 if it's not 8.: 1.0 \\
    replace the second element with the number 8, if it exists.: 1.0 \\
    replace the second element with 8 if it exists.: 1.0 \\
    replace the second element with 8.: 0.98 \\
    replace 1 with 8.: 0.38 \\
    replace the second element with the number 8 if the original second number is 1, 7, or 9.: 0.25 \\
    \textbf{Ground truth:} replace element 2 with an 8 if there is an element 2.
    \tcbseparator
    \textbf{Learned rule library:} \\
    swap the first and fourth elements.: 1.0 \\
    swap the first and fourth element of the list.: 1.0 \\
    swap the first three elements.: 1.0 \\
    swap the first and the fourth element of the list.: 1.0 \\
    swap the first and fourth element.: 1.0 \\
    \textbf{Ground truth:} swap elements 1 and 4.
    \tcbseparator
    \textbf{Learned rule library:} \\
    remove the first element.: 1.0 \\
    remove the first element from the list.: 1.0 \\
    remove the first element if the first element is 1 or 4.: 0.13 \\
    \textbf{Ground truth:} remove element 1.
    \tcbseparator
    \textbf{Learned rule library:} \\
    add 9 at the start and 7 at the end of the list.: 1.0 \\
    append 9 to the start of the list and 7 to the end of the list.: 1.0 \\
    add 9 to the start of the list and 7 to the end.: 1.0 \\
    add 9 at the start of the list and 7 at the end of the list.: 1.0 \\
    prepend number 9 and append number 7.: 1.0 \\
    append a 9 at the start of the list and a 7 at the end of the list.: 1.0 \\
    add 9 to the beginning and 7 to the end of the list.: 1.0 \\
    append 9 at the start of the list and 7 at end of the list.: 1.0 \\
    prepend the number 9 and append the number 7.: 1.0 \\
    add 9 at the beginning and 7 at the end of the list.: 1.0 \\
    add 9 at the beginning and 7 at the end.: 1.0 \\
    prepend 9 and append 7.: 1.0 \\
    add 9 at the start and 7 at the end of each list.: 1.0 \\
    add 9 to the front and 7 to the back of the list.: 1.0 \\
    add 9 at the start of the list and 7 at the end.: 1.0 \\
    \textbf{Ground truth:} prepend 9 and append 7.
\end{prompt}
\begin{prompt}[prm:list_functions_p2_rules]{75 rules learned by GPT-4 on 5 tasks selected from the P2 subset of List Functions. The P2 subset contains simple operations over numbers between 0 and 99.}
    \footnotesize
    \textbf{Learned rule library:} \\
    select the third element from the list.: 1.0 \\
    select the third element of the list.: 1.0 \\
    return a list with the third element from the given list.: 1.0 \\
    return the third element from the list.: 1.0 \\
    select the third element if there is one; otherwise, it selects none.: 1.0 \\
    retrieve the third element from the list.: 1.0 \\
    take the third element from the list.: 1.0 \\
    select the third value in the list.: 1.0 \\
    return the third element.: 1.0 \\
    return the third element in the list.: 1.0 \\
    select the 3rd element from the list.: 1.0 \\
    find the third element in the list.: 0.88 \\
    select the smallest odd number that is not 1.: 0.13 \\
    select the smallest odd number.: 0.13 \\
    \textbf{Ground truth:} remove all but element 3.
    \tcbseparator
    \textbf{Learned rule library:} \\
    return the last non-zero element.: 0.5 \\
    return a list with the maximum number that comes right after a number 1.: 0.5 \\
    return a list with the element that corresponds to the first number in the list, where indexing starts at 1.: 0.5 \\
    return the largest number that is divisible by 3.: 0.38 \\
    return a list containing the fourth element.: 0.38 \\
    select the maximum number that is less than or equal to 44.: 0.38 \\
    select the element directly after the digit 1.: 0.38 \\
    display the element after the number 1 or 2 in the array.: 0.38 \\
    extract the element next to 3, if the list contains a 3.: 0.25 \\
    return a list with the first number that ends with 4 or 5.: 0.25 \\
    find the first number that is divisible by both 2 and 5.: 0.13 \\
    return the second element.: 0.125 \\
    return the element that follows the number 1.: 0.13 \\
    select any number that follows a single digit number.: 0.13 \\
    return the third element.: 0.13 \\
    get the second highest number from the list.: 0.13 \\
    \textbf{Ground truth:} remove all but element N + 1, N = element 1.
    \tcbseparator
    \textbf{Learned rule library:} \\
    interchange the second and third elements.: 0.5 \\
    swap the third and second elements.: 0.5 \\
    swap the 2nd and 3rd element.: 0.5 \\
    swap the second and third elements.: 0.44 \\
    swap the second and third elements of the list.: 0.38 \\
    swap the first value with the fourth value if there is a fourth value in the list, and if not, the list remains the same.: 0.38 \\
    sort the first three items in ascending order, while keeping the rest of the items in their original order.: 0.25 \\
    place the smallest non-zero number on the list to the start of the list; if there are multiple instances of the smallest, it only moves the first encounter.: 0.13 \\
    \textbf{Ground truth:} swap elements 2 and 3 if element 2 > element 3, else swap elements 1 and 4.
    \tcbseparator
    \textbf{Learned rule library:} \\
    replace all elements with 10 copies of the first element.: 1.0 \\
    repeat the first number in the list 10 times.: 1.0 \\
    return a list with the first element of the input repeated 10 times.: 1.0 \\
    create a list of 10 elements which all have the same value as the first element in the input list.: 1.0 \\
    replace all elements with the first element repeated 10 times.: 1.0 \\
    replace all elements of the list with 10 instances of the first element.: 1.0 \\
    replace the element or elements of the list with ten times the first element of the list.: 1.0 \\
    replace all elements with ten times the first element.: 1.0 \\
    create a new list containing 10 copies of the first element of the original list.: 1.0 \\
    represent the first number in the list ten times.: 1.0 \\
    replace all elements in the list to the first element of the list and extend the list size to 10.: 1.0 \\
    replace all of the elements in the list with ten instances of the first element of the original list.: 1.0 \\
    generate a list of 10 repeats of the first element.: 1.0 \\
    repeat the first element of the list 10 times.: 1.0 \\
    replace all the elements of the list with the first element repeated 10 times.: 1.0 \\
    replace all elements with 10 instances of the first element.: 1.0 \\
    generate a list of ten instances of the first element.: 1.0 \\
    return a list of 10 elements all equal to the first element of the input.: 1.0 \\
    replace all elements by the first element and create a list of length 10.: 1.0 \\
    \textbf{Ground truth:} repeat element 1 ten times.
    \tcbseparator
    \textbf{Learned rule library:} \\
    prepend [11, 21, 43, 19] and append [7, 89, 0, 57].: 1.0 \\
    append the numbers 11, 21, 43, 19 at the start of the list and the numbers 7, 89, 0, 57 at the end of the list.: 1.0 \\
    add [11, 21, 43, 19] at the start of the list and [7, 89, 0, 57] at the end of the list.: 1.0 \\
    prepend the list [11, 21, 43, 19] to the original list, then append the list [7, 89, 0, 57] to the result.: 1.0 \\
    append the sequence 11, 21, 43, 19 to the start of the list, and the sequence 7, 89, 0, 57 to the end of the list.: 1.0 \\
    prepend [11, 21, 43, 19] to the input list and append [7, 89, 0, 57] to the end.: 1.0 \\
    prepend [11, 21, 43, 19] and append [7, 89, 0, 57] to the given array.: 1.0 \\
    add [11, 21, 43, 19] in front of the list and [7, 89, 0, 57] at the end of the list.: 1.0 \\
    add [11, 21, 43, 19] to the front of the array and [7, 89, 0, 57] to the end of array.: 1.0 \\
    add [11, 21, 43, 19] at the beginning, and [7, 89, 0, 57] at the end.: 1.0 \\
    prepend the list [11, 21, 43, 19] and append the list [7, 89, 0, 57].: 1.0 \\
    prepend the list [11, 21, 43, 19], and append the list [7, 89, 0, 57] to the input list.: 1.0 \\
    prepend the list with [11, 21, 43, 19] and append the list with [7, 89, 0, 57].: 1.0 \\
    prepend [11, 21, 43, 19] and append [7, 89, 0, 57] to the list.: 1.0 \\
    add [11, 21, 43, 19] to the start and [7, 89, 0, 57] to the end of the list.: 1.0 \\
    insert fixed numbers at the start [11, 21, 43, 19] and end [7, 89, 0, 57] of the list.: 1.0 \\
    prepend the list with [11, 21, 43, 19] and append it with [7, 89, 0, 57].: 1.0 \\
    add the list [11, 21, 43, 19] before the input list and [7, 89, 0, 57] after the input list.: 1.0 \\
    \textbf{Ground truth:} concatenate [11, 21, 43, 19], input, and [7, 89, 0, 57].
\end{prompt}
\begin{prompt}[prm:list_functions_p3_rules]{40 rules learned by GPT-4 on 5 tasks selected from the P3 subset of List Functions. The P3 subset contains difficult operations over numbers between 0 and 99.}
    \footnotesize
    \textbf{Learned rule library:} \\
    completely ignore the input and always output the same list [11, 19, 24, 33, 42, 5, 82, 0, 64, 9].: 1.0 \\
    replace the entire list with a predefined sequence, [11, 19, 24, 33, 42, 5, 82, 0, 64, 9]: 1.0 \\
    always produce an unchanging output, irrespective of the input.: 1.0 \\
    return a specific list of numbers ([11, 19, 24, 33, 42, 5, 82, 0, 64, 9]) regardless of the input.: 1.0 \\
    return a pre-determined list of numbers regardless of what the input is.: 1.0 \\
    replace any given list with the list [11, 19, 24, 33, 42, 5, 82, 0, 64, 9].: 1.0 \\
    replace any list (even an empty one) with [11, 19, 24, 33, 42, 5, 82, 0, 64, 9].: 1.0 \\
    replace any input list with the list [11, 19, 24, 33, 42, 5, 82, 0, 64, 9].: 1.0 \\
    output the same list [11, 19, 24, 33, 42, 5, 82, 0, 64, 9] regardless of the input.: 1.0 \\
    return the list [11, 19, 24, 33, 42, 5, 82, 0, 64, 9] regardless of the input.: 1.0 \\
    replace any input with a specific list: [11, 19, 24, 33, 42, 5, 82, 0, 64, 9].: 1.0 \\
    always return the list [11, 19, 24, 33, 42, 5, 82, 0, 64, 9] regardless of the input list.: 1.0 \\
    replace any list with [11, 19, 24, 33, 42, 5, 82, 0, 64, 9].: 1.0 \\
    return a predefined list regardless of the input.: 1.0 \\
    ignore the input and always produce the same output list [11, 19, 24, 33, 42, 5, 82, 0, 64, 9].: 1.0 \\
    completely ignore the input and always output the list [11, 19, 24, 33, 42, 5, 82, 0, 64, 9].: 1.0 \\
    replace any input with the list [11, 19, 24, 33, 42, 5, 82, 0, 64, 9].: 1.0 \\
    \textbf{Ground truth:} the list [11,19, 24, 33, 42, 5, 82, 0, 64, 9].
    \tcbseparator
    \textbf{Learned rule library:} \\
    remove the elements that are divisible by 2 (excluding the first and last element).: 0.38 \\
    remove elements greater than 50 and less than 100.: 0.38 \\
    remove elements greater than 50.: 0.38 \\
    remove any number from the list that ends with 3 or 5.: 0.38 \\
    remove the elements if they are greater than 50 and less than 80.: 0.25 \\
    remove the last element if it is an odd number,: 0.25 \\
    remove all elements from the list that are multiples of 3.: 0.25 \\
    keep only the numbers between 0 and 100 exclusive in the list.: 0.25 \\
    discard the numbers between 50 and 80 inclusive.: 0.25 \\
    eliminate the numbers that end with 9 from the list.: 0.25 \\
    remove all elements from the list that are divisible by 3.: 0.13 \\
    remove every third element from the list, starting count from 1.: 0.13 \\
    remove every second element from the list.: 0.13 \\
    remove numbers greater than or equal to 90 and less than or equal to 5.: 0.13 \\
    \textbf{Ground truth:} keep only elements whose tens digit is even.
    \tcbseparator
    \textbf{Learned rule library:} \\
    None of the learned rules has confidence greater than 0.1. \\
    \textbf{Ground truth:} replace each element, M, with M + the input length - M’s index.
    \tcbseparator
    \textbf{Learned rule library:} \\
    list each element as many times as its tens digit.: 0.5 \\
    replace numbers smaller than 10 with other numbers.: 0.25 \\
    repeat each number in the list the same number of times as the largest digit in that number.: 0.25 \\
    repeat the numbers divisible by 10 once and repeat numbers that are multiples of 5 based on the number of tens present in them.: 0.13 \\
    replace each element with the largest value in the list an amount of times equal to its last digit.: 0.13 \\
    replicate each element in the list by the number of its tens digit.: 0.13 \\
    repeat the last two digit second number as per the first one digit number and so on till it reaches a last one digit number and repeats next two digit number as per the last one digit number.: 0.13 \\
    \textbf{Ground truth:} repeat each element N times, where N is its tens digit, in order of appearance.
    \tcbseparator
    \textbf{Learned rule library:} \\
    sum all even numbers.: 1.0 \\
    add all elements divisible by 2.: 1.0 \\
    sum all elements divided by 2, then subtract all odd numbers and add all even numbers.: 0.13 \\
    sum the elements divisible by 3.: 0.13 \\
    \textbf{Ground truth:} sum of even elements.
\end{prompt}

\section{Full Experimental Results}
\label{app:experiment}

Table~\ref{tab:clutrr_text_full} provides the full results on the textual version of CLUTRR. Table~\ref{tab:clutrr_ablation} and \ref{tab:arithmetic_ablation} show the full results for the ablation studies on random rules and the XML tagging trick introduced in Appendix~\ref{app:implementation}. For the XML tagging trick, there are 2 levels for CLUTRR (first relation, second relation) and 3 levels for Arithmetic (carry, first addend, second addend). We consider tagging with different levels of hierarchy. Since XML tagging requires the rules to be sorted, we further consider a variant with unsorted (i.e.\ randomly ordered) rules and no hierarchy. We can see the XML tagging trick significantly boosts the performance of \method in Table~\ref{tab:clutrr_ablation} and \ref{tab:arithmetic_ablation}. This suggests that even when the good rules are given, retrieval is an important ability in deductive reasoning.

We manually analyze the predictions of CoT and CoT + \method on 100 test examples from CLUTRR and Arithmetic (base-16). We classify the predictions into five categories
\begin{itemize}[leftmargin=*]
    \item Correct: correct examples.
    \item Incorrect rules (only): error examples solely due to factually incorrect rules.
    \item Incorrect rules (and other): error examples due to both incorrect rules and other reasons.
    \item Retrieval: error examples solely due to the retrieval of undesired, but factually correct rules.
    \item Non-retrieval: error examples that do not fit into any of the above categories.
\end{itemize}

\begin{table}[!h]
    \centering
    \footnotesize
    \caption{Results on the textual CLUTRR with rules learned on the symbolic CLUTRR.}
    \label{tab:clutrr_text_full}
    \begin{adjustbox}{max width=\textwidth}
        \begin{tabular}{llcccccccccc}
            \toprule
            \bf{Model} & \bf{Prompt} & \bf{2 hops} & \bf{3 hops} & \bf{4 hops} & \bf{5 hops} & \bf{6 hops} & \bf{7 hops} & \bf{8 hops} & \bf{9 hops} & \bf{10 hops} & \bf{Average} \\
            \midrule
            \multirow{2}{*}{GPT-3.5}
            & 5-shot CoT & 12.5 & 11.1 & 12.9 & 36.0 & 6.3 & 19.0 & 30.0 & 0.0 & 16.0 & 16.0 \\
            & + HtT & 12.5 & 22.2 & 12.9 & 32.0 & 25.0 & 23.8 & 6.7 & 3.8 & 8.0 & 16.3 (+0.3) \\
            \midrule
            \multirow{2}{*}{GPT-4}
            & 5-shot CoT & 50.0 & 50.0 & 71.0 & 68.0 & 50.0 & 47.6 & 26.7 & 34.6 & 40.0 & 48.7 \\
            & + HtT & 100.0 & 55.6 & 77.4 & 72.0 & 75.0 & 38.1 & 23.3 & 42.3 & 48.0 & \bf{59.1 (+10.4)} \\
            \bottomrule
        \end{tabular}
    \end{adjustbox}
\end{table}

\begin{table}[!h]
    \centering
    \footnotesize
    \caption{Ablation studies on CLUTRR. All the entries are based on GPT-4.}
    \label{tab:clutrr_ablation}
    \begin{adjustbox}{max width=\textwidth}
        \begin{tabular}{lcccccccccc}
            \toprule
            \bf{Prompt} & \bf{2 hops} & \bf{3 hops} & \bf{4 hops} & \bf{5 hops} & \bf{6 hops} & \bf{7 hops} & \bf{8 hops} & \bf{9 hops} & \bf{10 hops} & \bf{Average} \\
            \midrule
            5-shot CoT & 50.0 & 55.6 & 71.0 & 80.0 & 50.0 & 52.4 & 30.0 & 46.2 & 20.0 & 50.6 \\
            \midrule
            + random rules & 0.0 & 16.7 & 3.2 & 20.0 & 6.3 & 23.8 & 3.3 & 11.5 & 4.0 & 9.9 (-40.7) \\
            \midrule
            + \method (unsorted) & 87.5 & 50.0 & 58.1 & 76.0 & 56.3 & 52.4 & 43.3 & 46.2 & 44.0 & 57.1 (+6.5) \\
            + \method (sorted) & 100.0 & 61.1 & 77.4 & 72.0 & 62.5 & 42.9 & 53.3 & 38.5 & 32.0 & 60.0 (+9.4) \\
            + \method (1 tag) & 100.0 & 61.1 & 74.2 & 76.0 & 62.5 & 42.9 & 53.3 & 30.8 & 36.0 & 59.6 (+9.0) \\
            + \method (2 tags) & 100.0 & 61.1 & 74.2 & 84.0 & 75.0 & 38.1 & 56.7 & 53.8 & 36.0 & \bf{64.3 (+13.7)} \\
            \bottomrule
        \end{tabular}
    \end{adjustbox}
\end{table}

\begin{table}[!h]
    \centering
    \footnotesize
    \caption{Ablation studies on Arithmetic. All the entries are based on GPT-4.}
    \label{tab:arithmetic_ablation}
    \begin{adjustbox}{max width=\textwidth}
        \begin{tabular}{lcccccccccc}
            \toprule
            \multirow{2}{*}{\bf{Prompt}} & \multicolumn{3}{c}{\bf{Base-16}} & \multicolumn{3}{c}{\bf{Base-11}} & \multicolumn{3}{c}{\bf{Base-9}} & \multirow{2}{*}{\bf{Average}}\\
            & \bf{2 digits} & \bf{3 digits} & \bf{4 digits} & \bf{2 digits} & \bf{3 digits} & \bf{4 digits} & \bf{2 digits} & \bf{3 digits} & \bf{4 digits} \\
            \midrule
            5-shot CoT & 83.3 & 71.1 & 61.5 & 52.8 & 47.4 & 46.2 & 75.0 & 36.8 & 42.3 & 57.4 \\
            \midrule
            + random rules & 19.4 & 18.4 & 0.0 & 30.6 & 28.9 & 15.4 & 44.4 & 21.1 & 34.6 & 23.7 (-33.7) \\
            \midrule
            + \method (unsorted) & 91.7 & 81.6 & 65.4 & 80.6 & 52.6 & 57.7 & 77.8 & 55.3 & 42.3 & 67.2 (+9.8) \\
            + \method (sorted) & 97.2 & 89.5 & 76.9 & 80.6 & 60.5 & 65.4 & 86.1 & 50.0 & 46.2 & 72.5 (+15.1) \\
            + \method (1 tag) & 91.7 & 89.5 & 84.6 & 80.6 & 60.5 & 46.2 & 88.9 & 65.8 & 65.4 & 74.8 (+17.4) \\
            + \method (2 tags) & 91.7 & 84.2 & 96.2 & 88.9 & 63.2 & 46.2 & 94.4 & 63.2 & 61.5 & 76.6 (+19.2) \\
            + \method (3 tags) & 100.0 & 94.7 & 84.6 & 88.9 & 71.1 & 46.2 & 86.1 & 68.4 & 65.4 & \bf{78.4 (+21.0)} \\
            \bottomrule
        \end{tabular}
    \end{adjustbox}
\end{table}

In the main paper, we aggregate the five categories into three coarse categories: correct, incorrect rules and others. Table~\ref{tab:error_case_full} demonstrates the full results of error cases on base-16 and CLUTRR. It is observed that \method significantly reduced incorrect rules versus CoT. On CLUTRR, \method has a moderate ratio of errors in the retrieval category. This is because either the required rule is missing in the library, or the LLM cannot retrieve the required rule. We believe such an issue can be mitigated by learning a more complete rule library and finetuning the LLM on retrieval problems.

\begin{table}[!h]
    \centering
    \footnotesize
    \caption{Statistics of error cases.}
    \label{tab:error_case_full}
    \begin{adjustbox}{max width=\textwidth}
        \begin{tabular}{llccccc}
            \toprule
            \multirow{2}{*}{\bf{Dataset}} & \multirow{2}{*}{\bf{Method}} & \multirow{2}{*}{\bf{Correct}} & \multicolumn{2}{c}{\bf{Incorrect Rules}} & \multicolumn{2}{c}{\bf{Others}} \\
            & & & \bf{Only} & \bf{And Others} & \bf{Retrieval} & \bf{Non-Retrieval} \\
            \midrule
            \multirow{2}{*}{CLUTRR}
            & CoT & 48\% & 27\% & 7\% & N/A & 18\% \\
            & CoT+\method & 57\% & 12\% & 10\% & 14\% & 4\% \\
            \midrule
            \multirow{2}{*}{Base-16}
            & CoT & 73\% & 19\% & 3\% & N/A & 5\% \\
            & CoT+\method & 93\% & 2\% & 1\% & 0\% & 4\% \\
            \bottomrule
        \end{tabular}
    \end{adjustbox}
\end{table}

\section{Experimental Results of Gemini Pro}
\label{app:gemini}

Here we further evaluate \method and the baselines using \texttt{gemini-1.0-pro}. We denote this model as Gemini Pro. The temperature of Gemini Pro is set to 0.9 following its default value.

Table~\ref{tab:clutrr_gemini} lists the results of Gemini Pro on CLUTRR. Our \method improves the performance of CoT and LtM prompting by a large margin of 10.2-19.2\% in the average accuracy. Table~\ref{tab:arithmetic_gemini} shows the results on Arithmetic datasets, where our method also significantly improves both few-shot prompting methods. The results on List Functions are provided in Table~\ref{tab:list_functions_gemini}. We can see that \method consistently improves both raw accuracy and task accuracy, similar to the observation on GPT models. By comparing Table~\ref{tab:clutrr_gemini}-\ref{tab:list_functions_gemini} with Table~\ref{tab:clutrr}-\ref{tab:list_functions}, we conclude that Gemini Pro is on par or slightly better than GPT-3.5, whereas GPT-4 has the best reasoning ability among these models.

\begin{table}[!h]
    \centering
    \footnotesize
    \caption{Results on the symbolic version of CLUTRR.}
    \label{tab:clutrr_gemini}
    \begin{adjustbox}{max width=\textwidth}
        \begin{tabular}{llcccccccccc}
            \toprule
            \bf{Model} & \bf{Prompt} & \bf{2 hops} & \bf{3 hops} & \bf{4 hops} & \bf{5 hops} & \bf{6 hops} & \bf{7 hops} & \bf{8 hops} & \bf{9 hops} & \bf{10 hops} & \bf{Average} \\
            \midrule
            & 0-shot CoT & 12.5 & 50.0 & 9.7 & 8.0 & 12.5 & 4.8 & 20.0 & 3.8 & 8.0 & 14.4 \\
            \cmidrule{2-12}
            & 5-shot CoT & 37.5 & 22.2 & 25.8 & 48.0 & 31.3 & 28.6 & 26.7 & 19.2 & 32.0 & 30.1 \\
            Gemini Pro & + HtT & 100.0 & 55.6 & 51.6 & 68.0 & 43.8 & 19.0 & 43.3 & 34.6 & 28.0 & \bf{49.3 (+19.2)} \\
            \cmidrule{2-12}
            & 5-shot LtM & 25.0 & 22.2 & 38.7 & 48.0 & 37.5 & 28.6 & 20.0 & 23.1 & 12.0 & 28.3 \\
            & + HtT & 87.5 & 44.4 & 41.9 & 64.0 & 31.3 & 23.8 & 30.0 & 15.4 & 8.0 & \bf{38.5 (+10.2)} \\
            \bottomrule
        \end{tabular}
    \end{adjustbox}
\end{table}

\begin{table}[!h]
    \centering
    \footnotesize
    \caption{Results on Arithmetic. Gemini Pro (5-shot CoT) has 71.8\% accuracy on base-10.}
    \label{tab:arithmetic_gemini}
    \begin{adjustbox}{max width=\textwidth}
        \begin{tabular}{llcccccccccc}
            \toprule
            \multirow{2}{*}{\bf{Model}} & \multirow{2}{*}{\bf{Prompt}} & \multicolumn{3}{c}{\bf{Base-16}} & \multicolumn{3}{c}{\bf{Base-11}} & \multicolumn{3}{c}{\bf{Base-9}} & \multirow{2}{*}{\bf{Average}}\\
            & & \bf{2 digits} & \bf{3 digits} & \bf{4 digits} & \bf{2 digits} & \bf{3 digits} & \bf{4 digits} & \bf{2 digits} & \bf{3 digits} & \bf{4 digits} \\
            \midrule
            & 0-shot CoT & 16.7 & 2.6 & 0.0 & 0.0 & 0.0 & 0.0 & 13.9 & 13.2 & 0.0 & 5.2 \\
            \cmidrule{2-12}
            & 5-shot CoT & 77.8 & 50.0 & 26.9 & 36.1 & 18.4 & 15.4 & 47.2 & 21.1 & 15.4 & 34.3 \\
            Gemini Pro & + HtT & 91.7 & 57.9 & 38.5 & 55.6 & 36.8 & 23.1 & 77.8 & 28.9 & 23.1 & \bf{48.2 (+13.9)} \\
            \cmidrule{2-12}
            & 5-shot LtM & 91.7 & 47.4 & 23.1 & 36.1 & 23.7 & 7.7 & 41.7 & 18.4 & 11.5 & 33.5 \\
            & + HtT & 61.1 & 44.7 & 26.9 & 47.2 & 26.3 & 34.6 & 66.7 & 36.8 & 15.4 & \bf{40.0 (+6.5)} \\
            \bottomrule
        \end{tabular}
    \end{adjustbox}
\end{table}

\begin{table}[!h]
    \centering
    \footnotesize
    \caption{Results on List Functions.}
    \label{tab:list_functions_gemini}
    \begin{adjustbox}{max width=\textwidth}
        \begin{tabular}{llcccccccc}
            \toprule
            \multirow{2}{*}{\bf{Model}} & \multirow{2}{*}{\bf{Prompt}} & \multicolumn{4}{c}{\bf{Raw Accuracy}} & \multicolumn{4}{c}{\bf{Task Accuracy}} \\
            & & \bf{P1} & \bf{P2} & \bf{P3} & \bf{Average} & \bf{P1} & \bf{P2} & \bf{P3} & \bf{Average} \\
            \midrule
            & 0-shot CoT & 27.7 & 33.4 & 19.3 & 26.8 & 15.0 & 25.0 & 6.0 & 15.3 \\
            \cmidrule{2-10}
            Gemini Pro & 4-shot CoT & 43.4 & 37.5 & 22.6 & 34.5 & 25.0 & 30.0 & 9.3 & 21.4 \\
            & + HtT & 55.5 & 54.4 & 28.6 & \bf{46.1 (+11.6)} & 33.8 & 40.0 & 12.0 & \bf{28.6 (+7.2)} \\
            \bottomrule
        \end{tabular}
    \end{adjustbox}
\end{table}

\vfill

\section{Examples of Model Outputs}

\subsection{Relational Reasoning}

\begin{prompt}[prm:clutrr_example]{Example outputs on CLUTRR. Errors are marked in red. Upper: CoT. Lower: CoT+\method.}
    \footnotesize
    \textbf{Context:} The relations on the path from Christine to Nicole are mother, son, sister, grandmother. \\
    \textbf{Question:} Nicole is Christine's what?
    \tcbseparator
    \textbf{Answer:} \\
    For mother's son, we have mother's son is brother. So the relations are reduced to brother, sister, grandmother. \\
    For brother's sister, we have brother's sister is sister. So the relations are reduced to sister, grandmother. \\
    For sister's grandmother, we have \textcolor{red}{sister's grandmother is great-grandmother.} So the relations are reduced to great-grandmother. \\
    Therefore, the answer is great-grandmother.
    \tcbseparator
    \textbf{Answer:} \\
    For mother's son, we retrieve <mother><son>mother's son is brother. So the relations are reduced to brother, sister, grandmother. \\
    For brother's sister, we retrieve <brother><sister>brother's sister is sister. So the relations are reduced to sister, grandmother. \\
    For sister's grandmother, we retrieve <sister><grandmother>sister's grandmother is grandmother. So the relations are reduced to grandmother. \\
    Therefore, the answer is grandmother.
\end{prompt}

\begin{prompt}[prm:text_example]{Example outputs on CLUTRR. Errors are marked in red. Upper: CoT. Lower: CoT+\method.}
    \footnotesize
    \textbf{Context:} The relations on the path from Sarah to Christine are husband, son, son, uncle, son, brother, brother, sister. \\
    \textbf{Question:} Christine is Sarah's what?
    \tcbseparator
    \textbf{Answer:} \\
    The path is quite complex, let's break it down: \\
    For husband's son, we have husband's son is son. So the relations become son, son, uncle, son, brother, brother, sister. \\
    For son's son, we have son's son is grandson. So the relations become grandson, uncle, son, brother, brother, sister. \\
    For grandson's uncle, we have \textcolor{red}{grandson's uncle is father or uncle.} We choose father, given that uncle will introduce a sibling relationship, which is not consistent with the remaining relationships. So the relations become father, son, brother, brother, sister. \\
    For father's son, we have father's son is brother. So the relations become brother, brother, brother, sister. \\
    For brother's brother, we have brother's brother is brother. So the relations become brother, brother, sister. \\
    For brother's brother, we still have brother's brother is also brother. So we are left with brother, sister. \\
    For brother's sister, we have brother's sister is sister. So the final relation is sister. \\
    Therefore, the answer is sister.
    \tcbseparator
    \textbf{Answer:} \\
    For husband's son, we retrieve <husband><son>husband's son is son. So the relations are reduced to son, son, uncle, son, brother, brother, sister. \\
    For son's son, we retrieve <son><son>son's son is grandson. So the relations are reduced to grandson, uncle, son, brother, brother, sister. \\
    For grandson's uncle, we retrieve <grandson><uncle>grandson's uncle is son. So the relations are reduced to son, son, brother, brother, sister. \\
    Reapplying son's son, we again have <son><son>son's son is grandson. So the relations are reduced to grandson, brother, brother, sister. \\
    For grandson's brother, we retrieve <grandson><brother>grandson's brother is grandson. So the relations are reduced to grandson, brother, sister. \\
    Reapplying grandson's brother, we have <grandson><brother>grandson's brother is grandson. So the relations are reduced to grandson, sister. \\
    For grandson's sister, we retrieve <grandson><sister>grandson's sister is granddaughter. So the relations are reduced to granddaughter. \\
    Therefore, the answer is granddaughter.
\end{prompt}

\subsection{Numerical Reasoning}

\begin{prompt}[prm:arithmetic_example1]{Example outputs on Arithmetic. Errors are marked in red. Upper: CoT. Lower: CoT+\method.}
    \footnotesize
    \textbf{Question:} In base-16, what is F55 + 85E?
    \tcbseparator
    \textbf{Answer:} \\
    F55 is F, 5, 5. 85E is 8, 5, E. So the steps are 5 + E, 5 + 5, F + 8. \\
    There is no carry. \textcolor{red}{5 + E = 19.} 19 is 1, 9. So we set the carry to 1. Prepend 9 to the answer. So far the answer has 1 digit: 9. \\
    The carry is 1. \textcolor{red}{5 + 5 + 1 = 11.} 11 is 1, 1. So we set the carry to 1. Prepend 1 to the answer. So far the answer has 2 digits: 1, 9. \\
    The carry is 1. \textcolor{red}{F + 8 + 1 = 1E.} \textcolor{red}{1E is 1, 4.} So we set the carry to 1. Prepend E to the answer. So far the answer has 3 digits: E, 1, 9. \\
    The carry is 1. Prepend 1 to the answer. So far the answer has 4 digits: 1, E, 1, 9. \\
    Therefore, the answer is 1E19.
    \tcbseparator
    \textbf{Answer:} \\
    F55 is F, 5, 5. 85E is 8, 5, E. So the steps are 5 + E, 5 + 5, F + 8. \\
    There is no carry. \textless no\_carry\textgreater \textless 5\textgreater \textless E\textgreater 5 + E = 13. 13 is 1, 3. So we set the carry to 1. Prepend 3 to the answer. So far the answer has 1 digit: 3. \\
    The carry is 1. \textless carry\textgreater \textless 5\textgreater \textless 5\textgreater 5 + 5 + 1 = B. B is 0, B. So we clear the carry. Prepend B to the answer. So far the answer has 2 digits: B, 3. \\
    There is no carry. \textless no\_carry\textgreater \textless F\textgreater \textless 8\textgreater F + 8 = 17. 17 is 1, 7. So we set the carry to 1. Prepend 7 to the answer. So far the answer has 3 digits: 7, B, 3. \\
    The carry is 1. Prepend 1 to the answer. So far the answer has 4 digits: 1, 7, B, 3. \\
    Therefore, the answer is 17B3. 
\end{prompt}

\begin{prompt}[prm:arithmetic_example2]{Example outputs on Arithmetic. Errors are marked in red. Upper: CoT. Lower: CoT+\method.}
    \footnotesize
    \textbf{Question:} In base-16, what is 47D2 + D292?
    \tcbseparator
    \textbf{Answer:} \\
    47D2 is 4, 7, D, 2. D292 is D, 2, 9, 2. So the steps are 2 + 2, D + 9, 7 + 2, 4 + D. \\
    There is no carry. 2 + 2 = 4. 4 is 0, 4. So we clear the carry. Prepend 4 to the answer. So far the answer has 1 digit: 4. \\
    There is no carry. \textcolor{red}{D + 9 = 1C.} 1C is 1, C. So we set the carry to 1. Prepend C to the answer. So far the answer has 2 digits: C, 4. \\
    The carry is 1. 7 + 2 + 1 = A. A is 0, A. So we clear the carry. Prepend A to the answer. So far the answer has 3 digits: A, C, 4. \\
    The carry is 0. 4 + D = 11. 11 is 1, 1. So we set the carry to 1. Prepend 1 to the answer. So far the answer has 4 digits: 1, A, C, 4. \\
    The carry is 1. Prepend 1 to the answer. So far the answer has 5 digits: 1, 1, A, C, 4. \\
    Therefore, the answer is 11AC4.
    \tcbseparator
    \textbf{Answer:} \\
    47D2 is 4, 7, D, 2. D292 is D, 2, 9, 2. So the steps are 2 + 2, D + 9, 7 + 2, 4 + D. \\
    There is no carry. \textless no\_carry\textgreater \textless 2\textgreater \textless 2\textgreater 2 + 2 = 4. 4 is 0, 4. So we clear the carry. Prepend 4 to the answer. So far the answer has 1 digit: 4. \\
    There is no carry. \textless no\_carry\textgreater \textless D\textgreater \textless 9\textgreater D + 9 = 16. 16 is 1, 6. So we set the carry to 1. Prepend 6 to the answer. So far the answer has 2 digits: 6, 4. \\
    The carry is 1. \textless carry\textgreater \textless 7\textgreater \textless 2\textgreater 7 + 2 + 1 = A. A is 0, A. So we clear the carry. Prepend A to the answer. So far the answer has 3 digits: A, 6, 4. \\
    There is no carry. \textless no\_carry\textgreater \textless 4\textgreater \textless D\textgreater 4 + D = 11. 11 is 1, 1. So we set the carry to 1. Prepend 1 to the answer. So far the answer has 4 digits: 1, A, 6, 4. \\
    The carry is 1. Prepend 1 to the answer. So far the answer has 5 digits: 1, 1, A, 6, 4. \\
    Therefore, the answer is 11A64.
\end{prompt}

\subsection{Concept Learning}

\begin{prompt}[prm:list_functions_example1]{Example outputs on List Functions. Errors are marked in red. Upper: CoT. Lower: CoT+\method.}
    \footnotesize
    \textbf{Examples:} \\[0pt]
    [20] -> [20] \\[0pt]
    [83, 8, 89, 63, 12, 4, 76, 7] -> [4, 8, 12, 16, 20, 24, 28, 32] \\[0pt]
    [8, 6, 36] -> [6, 12, 18] \\[0pt]
    [22, 25] -> [22, 44] \\[0pt]
    [19] -> [19] \\[0pt]
    [14, 17] -> [14, 28] \\[0pt]
    [57, 82, 23] -> [23, 46, 69] \\[0pt]
    [84] -> [84] \\[0pt]
    [] -> [] \\[0pt]
    [44] -> [44] \\[0pt]
    [67, 46, 27] -> [27, 54, 81] \\[0pt]
    [58, 33] -> [33, 66] \\[0pt]
    [92, 71, 71, 5] -> [5, 10, 15, 20] \\[0pt]
    [46, 46] -> [46, 92] \\[0pt]
    [47] -> [47] \\[0pt]
    [56, 97, 24, 79] -> [24, 48, 72, 96] \\
    \textbf{Questions:} \\[0pt]
    [78, 41] -> ? \\[0pt]
    [91] -> ? \\[0pt]
    [53] -> ? \\[0pt]
    [7, 44, 98] -> ? \\[0pt]
    [37] -> ? \\[0pt]
    [59] -> ? \\[0pt]
    [13, 24, 13] -> ? \\[0pt]
    [69, 93, 12, 84] -> ? \\[0pt]
    [9, 96, 15, 56] -> ? \\[0pt]
    [31, 60, 47] -> ? \\[0pt]
    [8, 86, 35, 22, 9] -> ? \\[0pt]
    [46, 7, 9, 2, 14, 4, 80, 72, 54] -> ? \\[0pt]
    [86, 28] -> ? \\[0pt]
    [2, 23, 94, 91, 14, 1, 36, 32, 97, 77] -> ? \\[0pt]
    [6, 4, 37, 3, 60, 8, 9] -> ? \\[0pt]
    [68] -> ?
    \tcbseparator
    \textbf{Answers:} \\
    From the examples, we infer the function is to generate a sequence of multiples of \textcolor{red}{the last element} in the list. \\
    Using this function, the answers to the questions are: \\[0pt]
    [78, 41] -> [41, 82] \\[0pt]
    [91] -> [91] \\[0pt]
    [53] -> [53] \\[0pt]
    [7, 44, 98] -> \textcolor{red}{[98, 196, 294]} \\[0pt]
    [37] -> [37] \\[0pt]
    [59] -> [59] \\[0pt]
    [13, 24, 13] -> [13, 26, 39] \\[0pt]
    [69, 93, 12, 84] -> \textcolor{red}{[84, 168, 252, 336]} \\[0pt]
    [9, 96, 15, 56] -> \textcolor{red}{[56, 112, 168, 224]} \\[0pt]
    [31, 60, 47] -> \textcolor{red}{[47, 94, 141]} \\[0pt]
    [8, 86, 35, 22, 9] -> \textcolor{red}{[9, 18, 27, 36, 45]} \\[0pt]
    [46, 7, 9, 2, 14, 4, 80, 72, 54] -> \textcolor{red}{[54, 108, 162, 216, 270, 324, 378, 432, 486]} \\[0pt]
    [86, 28] -> [28, 56] \\[0pt]
    [2, 23, 94, 91, 14, 1, 36, 32, 97, 77] -> \textcolor{red}{[77, 154, 231, 308, 385, 462, 539, 616, 693, 770]} \\[0pt]
    [6, 4, 37, 3, 60, 8, 9] -> \textcolor{red}{[9, 18, 27, 36, 45, 54, 63]} \\[0pt]
    [68] -> [68]
    \tcbseparator
    \textbf{Answers:} \\
    Based on the examples and the potential functions, we infer the function is to take the smallest number in the list and multiply it by i+1 for the new list, where i is the index. \\
    Using this function, the answers to the questions are: \\[0pt]
    [78, 41] -> [41, 82] \\[0pt]
    [91] -> [91] \\[0pt]
    [53] -> [53] \\[0pt]
    [7, 44, 98] -> [7, 14, 21] \\[0pt]
    [37] -> [37] \\[0pt]
    [59] -> [59] \\[0pt]
    [13, 24, 13] -> [13, 26, 39] \\[0pt]
    [69, 93, 12, 84] -> [12, 24, 36, 48] \\[0pt]
    [9, 96, 15, 56] -> [9, 18, 27, 36] \\[0pt]
    [31, 60, 47] -> [31, 62, 93] \\[0pt]
    [8, 86, 35, 22, 9] -> [8, 16, 24, 32, 40] \\[0pt]
    [46, 7, 9, 2, 14, 4, 80, 72, 54] -> [2, 4, 6, 8, 10, 12, 14, 16, 18] \\[0pt]
    [86, 28] -> [28, 56] \\[0pt]
    [2, 23, 94, 91, 14, 1, 36, 32, 97, 77] -> [1, 2, 3, 4, 5, 6, 7, 8, 9, 10] \\[0pt]
    [6, 4, 37, 3, 60, 8, 9] -> [3, 6, 9, 12, 15, 18, 21] \\[0pt]
    [68] -> [68]
\end{prompt}

\begin{prompt}[prm:list_functions_example2]{Example outputs on List Functions. Errors are marked in red. Upper: CoT. Lower: CoT+\method.}
    \footnotesize
    \textbf{Examples:} \\[0pt]
    [23, 3, 6, 8, 67, 9, 2, 0, 4, 46] -> [23, 99, 46] \\[0pt]
    [7, 2, 8, 4, 0, 32, 4] -> [7, 46, 4] \\[0pt]
    [6, 22, 1, 7, 20, 14, 14, 14] -> [6, 78, 14] \\[0pt]
    [9, 1, 3, 7, 4, 36] -> [9, 15, 36] \\[0pt]
    [31, 7, 1, 62, 2, 7, 8, 6, 4] -> [31, 93, 4] \\[0pt]
    [5, 7, 19, 6, 2, 0, 1] -> [5, 34, 1] \\[0pt]
    [8, 3, 1, 0, 5, 66] -> [8, 9, 66] \\[0pt]
    [30, 11, 41, 0, 9, 6] -> [30, 61, 6] \\[0pt]
    [0, 35, 1, 9, 6, 4, 3, 56] -> [0, 58, 56] \\[0pt]
    [4, 0, 5, 28, 28, 28, 4] -> [4, 89, 4] \\[0pt]
    [6, 0, 7, 25, 33, 4, 18, 7] -> [6, 87, 7] \\[0pt]
    [2, 3, 7, 5, 5, 5, 5] -> [2, 25, 5] \\[0pt]
    [26, 37, 6, 5, 3, 4] -> [26, 51, 4] \\[0pt]
    [10, 9, 3, 8, 5, 1, 6] -> [10, 26, 6] \\[0pt]
    [60, 11, 63, 4, 11, 1, 7] -> [60, 90, 7] \\[0pt]
    [0, 29, 26, 7, 29, 8] -> [0, 91, 8] \\
    \textbf{Questions:} \\[0pt]
    [55, 7, 24, 6, 1, 0, 3, 4] -> ?\\[0pt]
    [5, 8, 32, 6, 2, 0] -> ?\\[0pt]
    [84, 70, 9, 0, 1, 32] -> ?\\[0pt]
    [90, 19, 8, 1, 7, 0, 5] -> ?\\[0pt]
    [4, 2, 0, 4, 6, 0, 2, 2, 6, 0] -> ?\\[0pt]
    [76, 1, 44, 47, 7, 8] -> ?\\[0pt]
    [1, 5, 14, 9, 8, 3, 6, 88] -> ?\\[0pt]
    [58, 1, 3, 12, 2, 5, 8] -> ?\\[0pt]
    [96, 4, 0, 1, 63, 17] -> ?\\[0pt]
    [9, 0, 24, 4, 50, 0, 7] -> ?\\[0pt]
    [46, 4, 16, 7, 1, 43, 1, 5, 1, 43] -> ?\\[0pt]
    [74, 9, 3, 0, 44, 54] -> ?\\[0pt]
    [8, 81, 7, 1, 3, 0] -> ?\\[0pt]
    [33, 7, 8, 24, 29, 21, 1, 5] -> ?\\[0pt]
    [0, 2, 30, 30, 1, 6, 4, 8] -> ?\\[0pt]
    [8, 1, 8, 15, 4, 2, 0, 0, 40] -> ?
    \tcbseparator
    \textbf{Answers:} \\
    From the examples, we infer the function is to keep the first and last elements, and \textcolor{red}{replace the second element with the sum of the second and third elements.} \\
    Using this function, the answers to the questions are: \\[0pt]
    [55, 7, 24, 6, 1, 0, 3, 4] -> [55, \textcolor{red}{31}, 4] \\[0pt]
    [5, 8, 32, 6, 2, 0] -> [5, \textcolor{red}{40}, 0] \\[0pt]
    [84, 70, 9, 0, 1, 32] -> [84, \textcolor{red}{79}, 32] \\[0pt]
    [90, 19, 8, 1, 7, 0, 5] -> [90, \textcolor{red}{27}, 5] \\[0pt]
    [4, 2, 0, 4, 6, 0, 2, 2, 6, 0] -> [4, \textcolor{red}{2}, 0] \\[0pt]
    [76, 1, 44, 47, 7, 8] -> [76, \textcolor{red}{45}, 8] \\[0pt]
    [1, 5, 14, 9, 8, 3, 6, 88] -> [1, \textcolor{red}{19}, 88] \\[0pt]
    [58, 1, 3, 12, 2, 5, 8] -> [58, \textcolor{red}{4}, 8] \\[0pt]
    [96, 4, 0, 1, 63, 17] -> [96, \textcolor{red}{4}, 17] \\[0pt]
    [9, 0, 24, 4, 50, 0, 7] -> [9, \textcolor{red}{24}, 7] \\[0pt]
    [46, 4, 16, 7, 1, 43, 1, 5, 1, 43] -> [46, \textcolor{red}{20}, 43] \\[0pt]
    [74, 9, 3, 0, 44, 54] -> [74, \textcolor{red}{12}, 54] \\[0pt]
    [8, 81, 7, 1, 3, 0] -> [8, \textcolor{red}{88}, 0] \\[0pt]
    [33, 7, 8, 24, 29, 21, 1, 5] -> [33, \textcolor{red}{15}, 5] \\[0pt]
    [0, 2, 30, 30, 1, 6, 4, 8] -> [0, \textcolor{red}{32}, 8] \\[0pt]
    [8, 1, 8, 15, 4, 2, 0, 0, 40] -> [8, \textcolor{red}{9}, 40]
    \tcbseparator
    \textbf{Answers:} \\
    Based on the examples and the potential functions, we infer the function is to keep the first number, then the sum of the middle numbers, then the last number in the list. \\
    Using this function, the answers to the questions are: \\[0pt]
    [55, 7, 24, 6, 1, 0, 3, 4] -> [55, 41, 4] \\[0pt]
    [5, 8, 32, 6, 2, 0] -> [5, 48, 0] \\[0pt]
    [84, 70, 9, 0, 1, 32] -> [84, 80, 32] \\[0pt]
    [90, 19, 8, 1, 7, 0, 5] -> [90, 35, 5] \\[0pt]
    [4, 2, 0, 4, 6, 0, 2, 2, 6, 0] -> [4, 22, 0] \\[0pt]
    [76, 1, 44, 47, 7, 8] -> [76, 99, 8] \\[0pt]
    [1, 5, 14, 9, 8, 3, 6, 88] -> [1, 45, 88] \\[0pt]
    [58, 1, 3, 12, 2, 5, 8] -> [58, 23, 8] \\[0pt]
    [96, 4, 0, 1, 63, 17] -> [96, 68, 17] \\[0pt]
    [9, 0, 24, 4, 50, 0, 7] -> [9, 78, 7] \\[0pt]
    [46, 4, 16, 7, 1, 43, 1, 5, 1, 43] -> [46, \textcolor{red}{77}, 43] \\[0pt]
    [74, 9, 3, 0, 44, 54] -> [74, 56, 54] \\[0pt]
    [8, 81, 7, 1, 3, 0] -> [8, 92, 0] \\[0pt]
    [33, 7, 8, 24, 29, 21, 1, 5] -> [33, 90, 5] \\[0pt]
    [0, 2, 30, 30, 1, 6, 4, 8] -> [0, 73, 8] \\[0pt]
    [8, 1, 8, 15, 4, 2, 0, 0, 40] -> [8, 30, 40] \\[0pt]
\end{prompt}

\end{document}